\newif\ifelspaper

\newif\ifarxiv 
\ifelspaper
\documentclass[final,5p,times,twocolumn]{elsarticle}
\else
\arxivtrue
\documentclass[preprint,12pt]{elsarticle}
\fi

\usepackage{amssymb}

\usepackage[utf8]{inputenc}
\usepackage{graphicx}
\ifarxiv
\usepackage{hyperref}
\else
\usepackage[colorlinks]{hyperref}
\fi
\usepackage[nolist]{acronym}
\usepackage{subcaption}
\usepackage{siunitx}                         
\usepackage{color}
\usepackage{amsmath}
\usepackage{xspace}
\newcommand{\nederdrone}{\emph{NederDrone}\xspace}
\newcommand{\fig}{Fig.}
\ifelspaper
\newcommand{\suffix}{_els}
\else
\newcommand{\suffix}{}
\fi

\newcommand{\myurl}[1]{\href{#1}{#1}}

\date{November 2020}

\ifelspaper
\journal{International Journal of Hydrogen Energy}
\fi

\begin{document}

\begin{frontmatter}


\title{The \nederdrone: a hybrid lift, hybrid energy\\hydrogen UAV}


\author[tudelft]{C. De Wagter}
\ifarxiv
\else
\ead{c.dewagter@tudelft.nl}
\fi

\author[tudelft]{B. Remes}
\author[tudelft]{E. Smeur}
\author[tudelft]{F. van Tienen}
\author[tudelft]{\\R. Ruijsink}
\author[tudelft]{K. van Hecke}
\author[tudelft]{E. van der Horst}

\address[tudelft]{Micro Air Vehicle Lab\fnref{mavlab}, TUDelft, Kluyverweg 1, 2629HS Delft, the Netherlands}

\begin{abstract}
\ifelspaper
\else

A lot of UAV applications require vertical take-off and landing (VTOL) combined with very long-range or endurance.
Transitioning UAVs have been proposed to combine the VTOL capabilities of helicopters with the efficient long-range flight properties of fixed-wing aircraft.
But energy is still a bottleneck for many electric long endurance applications.
While solar power technology and battery technology have improved a lot, in rougher conditions they still respectively lack the power or total amount of energy required for many real-world situations.
In this paper, we introduce the \nederdrone, a hybrid lift, hybrid energy hydrogen-powered UAV which is able to perform vertical take-off and landings using 12 propellers while flying efficiently in forward flight thanks to its fixed wings.
The energy is supplied from a mix of 
hydrogen-driven fuel-cells to store large amounts of energy and
battery power for high power situations.
The hydrogen is stored in a pressurized cylinder around which the UAV is optimized.
This paper analyses the selection of the concept, the implemented safety elements, the electronics and flight control and shows flight data including a 3h38 flight at sea, starting and landing on a small moving ship.

\fi
\end{abstract}


\end{frontmatter}


\section{Introduction}

\begin{acronym}

\acro{CAN}{Controller Area Network}
\acro{CNG}{Compressed Natural Gas}
\acro{CTS}{Composite Technical Systems}
\acro{ESC}{Electronic Speed Controller}
\acro{EPP}{Expanded Polypropylene}
\acro{INDI}{Incremental Nonlinear Dynamic Inversion}
\acro{LHV}{Lower Heating Value}
\acro{OJF}{Open Jet Facility}
\acro{PCP}{Pre-Charged Pneumatic}
\acro{PEM}{Polymer Electrolyte Membrane}
\acro{PET}{Polyethylene Terephthalate}
\acro{TPED}{Transportable Pressure Equipment Directive}
\acro{UAV}{Unmanned Air Vehicles}
\acro{VTOL}{Vertical Take-Off and Landing}
\end{acronym}

\ac{UAV} offer solutions in a large variety of applications \citep{Pajares2015_overview}.
While a lot of applications can be performed with current battery technology, for many others the energy requirements cannot be met \citep{Boukoberine2019_UAV_energy}.
In particular when combined with the requirement to have \ac{VTOL} capabilities, the traditional efficient fixed-wing aircraft is not an option.
This is where a lot of hybrid concepts, namely combinations of efficient fixed-wings and hovering rotorcraft, have been proposed in recent years.

\begin{figure*}[htb]
\begin{subfigure}{.32\textwidth}
  \centering
  \includegraphics[width=.95\linewidth]{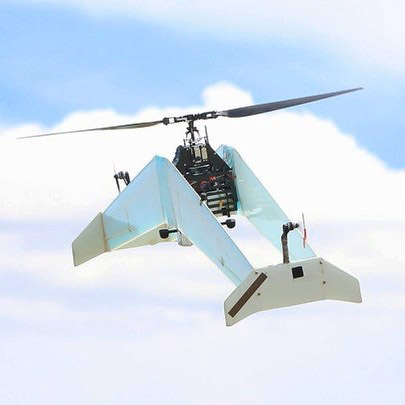}  
  \caption{DelftaCopter \citep{wagter_2018_delftacopter}}
  \label{fig:hybrid:delftacopter}
\end{subfigure}
\begin{subfigure}{.32\textwidth}
  \centering
  \includegraphics[width=.95\linewidth]{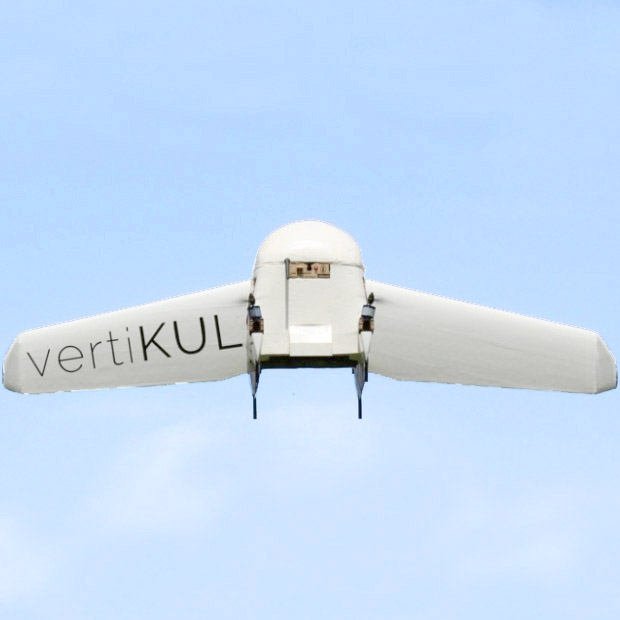}  
  \caption{VertiKUL \citep{Hochstenbach2015_vertikul}}
  \label{fig:hybrid:vertikul}
\end{subfigure}
\begin{subfigure}{.32\textwidth}
  \centering
  \includegraphics[width=.95\linewidth]{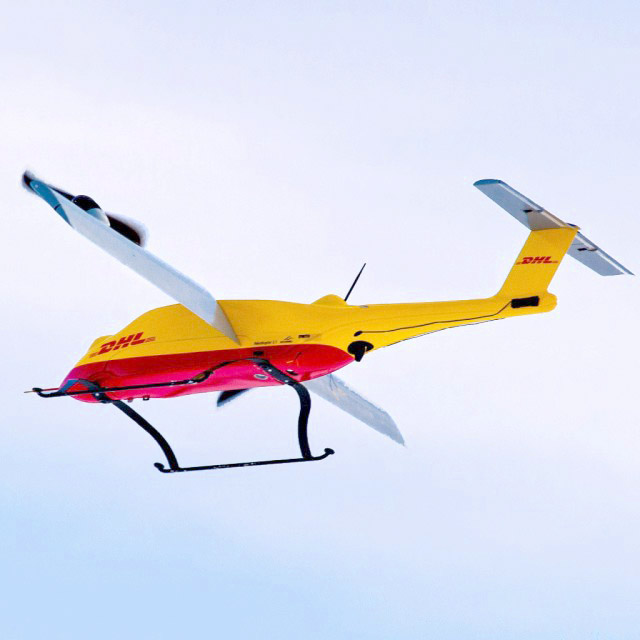}  
  \caption{
  DHL Parcelcopter 3
  \citep{Hartmann2017_tiltwing}}
  \label{fig:hybrid:tiltwing}
\end{subfigure}
\caption{Hybrid UAV able to take-off and land vertically while using fixed-wings to increase flight efficiency in forward flight. }
\label{fig:hybrid}
\end{figure*}

\subsection{Hybrid Lift}\label{sec:hybrid_lift}

The most common categories of hybrid lift \ac{UAV} are the tail-sitters, dual-systems and transforming \ac{UAV} \cite{Saeed2018_hybrid_cats}.
Tail-sitters pitch down \ang{90} during the transition from hover to forward flight, and while they have important draw-backs for pilot comfort \citep{Anderson1981_vtol}, they have gained a lot of new interest for \ac{UAV}.
They are mechanically simple yet allow to re-use propulsion systems in several phases of the flight \citep{Hochstenbach2015_vertikul}.
Many different types of tail-sitters exist. They can either be optimized to maximize the hovering efficiency with a single large rotor \citep{wagter_2018_delftacopter} (See \fig\ref{fig:hybrid:delftacopter}) or to minimize complexity \citep{Smeur2019_cyclone_indi}.
Other tail-sitters were optimized for maximal redundancy \citep{wagter_2020_nederdrone1} or were given re-configurable wings to minimize sensitivity to gusts in hover \cite{Wang2017_convert_tailsitter}.

The second category is formed by dual-systems like quad-planes.
These \ac{UAV} contain a complete hovering vehicle in-plane with a separate fixed-wing vehicle. Both parts are typically operated separately \citep{Govdeli2019_quadplane}.

The last category are transforming vehicles which try to re-use propulsion systems in hover and forward flight by either tilting the entire wings with respect to the fuselage \citep{Hartmann2017_tiltwing} (See \fig\ref{fig:hybrid:tiltwing}), or by only tilting the motors \citep{Flores2011_tiltrotor}.

\subsection{Hydrogen UAV}\label{sec:hyrdogen}

But despite large advancements in battery technology and drone technology, energy storage has remained the biggest bottleneck for the endurance of \ac{UAV}.
Recent advancement in lightweight robust fuel-cell technology has increased the interest of using them in UAV applications \citep{Pan2019_recent_fuelcell_advancements}.

Fuel-cells for \ac{UAV} have been proposed for a long time \citep{Furrutter2009_small_fuelcell_100W}.
But a lot of problems had to be overcome before getting small reliable membranes which can power more than just the most efficient \ac{UAV} \citep{Gong2017_propulsion}. 
In several \ac{UAV}s the fuel-cell power had to be complemented with for instance lithium-ion batteries \citep{Herwerth2007_hydro_uav_150W}.
In the last decade, hydrogen fuel-cell powered quadrotors have been developed \citep{Kim2012_hydrogen_quad}, which show the viability of the concept.
But flight times of hydrogen powered multi-copters never reach the endurance seen in fixed-wings.

Many fixed wing hydrogen UAVs have been proposed like the \SI{16}{\kilogram} \SI{500}{\watt} demonstrator from \citep{Bradley2007_FX_500W} in 2007, the \SI{1.5}{\kilogram} \SI{100}{\watt} \ac{UAV} from \citep{Kim2012_hydrogen_quad} in 2012, the \SI{11}{\kilogram} \SI{200}{\watt} from \citep{LapenaRey2017_200W_FW} in 2017 to the 2020 \SI{6.4}{\kilogram} \SI{250}{\watt} \citep{Oezbek2020_250W_FW}.
Other projects investigated the combination of hydrogen power with solar power \citep{GADALLA2016_fuelcell_solar}, which effectively helped to double their flight time in ideal conditions. This combination has also been proposed to cross the Atlantic \citep{Gavritovic2019_atlantic}.
But in most projects, a combination of battery power for high demand situations and hydrogen power for endurance has been used, which is referred to as hybrid energy \citep{Rottmayer2011_fuelcell_naval}.

\subsection{Hybrid Lift Hybrid Energy}

To combine the advantages of hybrid lift \ac{UAV} with those of hybrid energy from batteries and hydrogen fuel-cells, a new concept is developed.
Section~\ref{sec:cylinder} investigates the selected type of fuel-cell and safety aspects of flying with hydrogen.
Section~\ref{sec:concept} explains the design choices of the hybrid \ac{UAV} built around the fuel-cell system.
Section~\ref{sec:electronics} explains the hybrid power wiring and dual control bus of the \nederdrone.
Section~\ref{sec:windtunnel} describes the essential aerodynamic properties.
Section~\ref{sec:control} explains the control.
Section~\ref{sec:flight} shows actual test flight data. Finally Section~\ref{sec:discussion} and \ref{sec:conclusion} give a discussion and conclusion respectively.


\section{Hydrogen powered electric flight}\label{sec:cylinder}

Hydrogen powered fuel-cells form an attractive solution for sustainable aircraft if the remaining technological problems can be solved \citep{Stephens2016_sustainable_FC}.
The main elements consist of a fuel-cell and backup power which are sufficient for the drone, a hydrogen storage solution and important safety considerations.

\subsection{Fuel-cell}

\begin{figure}[htb]
\centering
\includegraphics[width=0.95\linewidth]{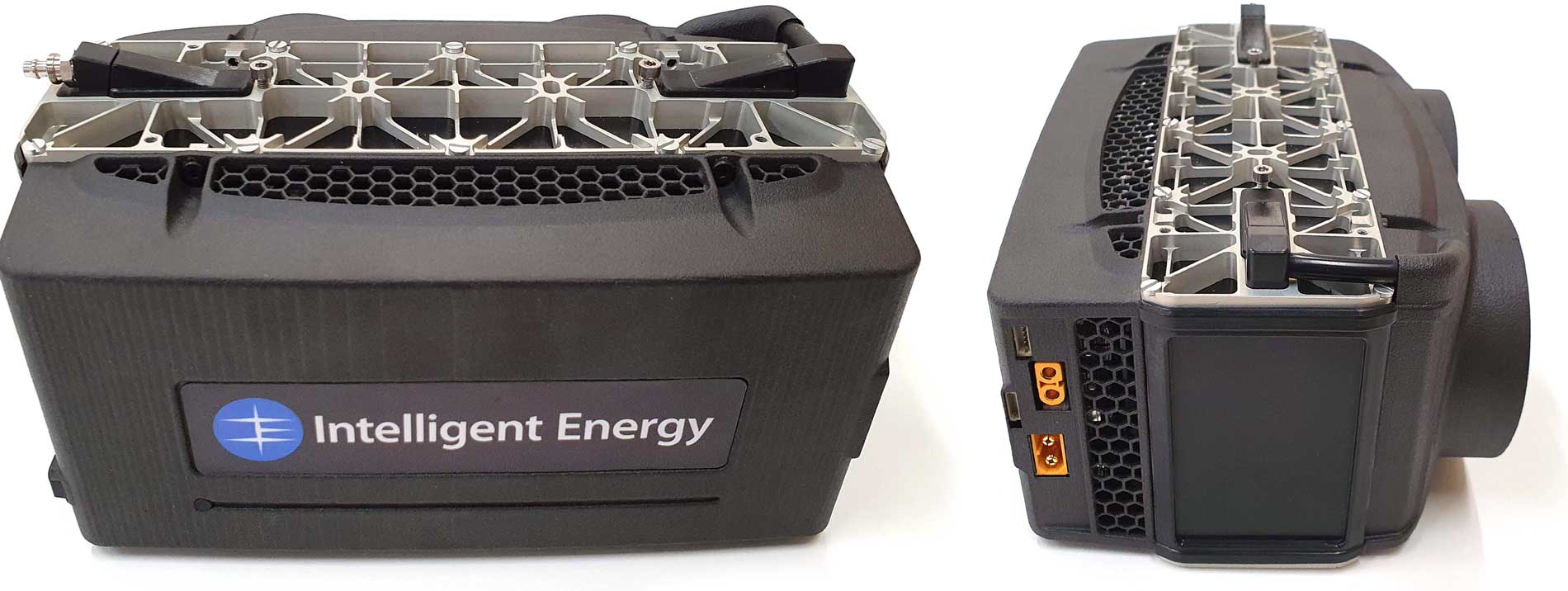}
\caption{Intelligent Energy \SI{800}{\watt} fuel-cell system.} \label{fig:intelligent_energy}
\end{figure}

The three most common fuel-cells used to power UAVs are: 1) \ac{PEM} fuel-cells, 2) direct methanol fuel-cells, and 3) solid oxide fuel cells \citep{Wang2020_fuel_cell_overview}.
But the availability of ready to use systems at the time of selection also plays an important role.
Although \ac{PEM} fuel-cell efficiency drops with altitude \citep{Horde2012_pem_altitude}, 
and their membrane must be re-humidified to unlock their full power when not used for a few days \citep{Kim2014_humidify},
they form an attractive choice for \ac{UAV}.
Two options within the power range from \SI{300}{\watt} to \SI{1000}{\watt} were available, namely \ac{PEM} fuel-cell systems from
Intelligent Energy\footnote{\myurl{http://www.intelligent-energy.com}} (\textbf{IE}) and 
HES Energy Systems\footnote{\myurl{https://www.hes.sg/}} (\textbf{HES}).

\begin{table}[hb]
\centering
\begin{tabular}{|lllllll|}
\hline
	& $P$	 & $P_{max}$ &	Lipo	& $\zeta$  & $W$ 	& $W_{p.r}$ \\ 
Unit	& [W]     &  [W]    &  [cell]  & [\%] & [kg] & [kg] \\ \hline
\textbf{HES} & 250 & 250	&	6	&50\%	&0.73&	0.14\\
	& 500 & 500	&	7	&52\%	&1.4	&0.14\\
\textbf{IE}	&650	&1000	&6	&56\%	&0.81&	0.14\\
&	800	&1400	&6	&55\%	&0.96	&0.14\\ \hline
\end{tabular}
\caption{Available fuel-fell power $P$, system maximum power $P_{max}$, number of lithium cells, efficiency $\zeta$ and pressure reduced weight $W_{p.r}$.}
\end{table}

The \textbf{IE} \SI{800}{\watt} air-cooled \ac{PEM} fuel-cell running at ambient temperatures was selected (See \fig\ref{fig:intelligent_energy}), which is packaged as a small light-weight cost effective and robust system.
It runs at the easily available 6-cell lithium output voltage and---at the time of selection---had a better hydrogen efficiency and weight efficiency.

The \ac{LHV} efficiency $Eff_{PEMFC}$  of the \SI{800}{\watt} system is between \SI{53}{\percent} at \SI{800}{W} and \SI{56}{\percent} at \SI{700}{\watt}\footnote{IE: \href{https://www.intelligent-energy.com/uploads/product\_docs/61126\_IE\_-\_Cylinder\_Guide\_May\_2020.pdf}{uploads/product\_docs/61126\_IE\_-\_Cylinder\_Guide\_May\_2020.pdf}}.
The fuel consumption $ff_{H_2}$ in \SI{}{\gram/\hour} at predicted forward flight conditions of \SI{600}{\watt} average power use then becomes:

\begin{equation}
ff_{H_2} = \frac{P_{mean}}{E_{specific_{H_2}} \cdot Eff_{PEMFC}}\label{eq:specific}
\end{equation}

A hydrogen \ac{LHV} of \SI{33.3}{\watt.\hour/\gram} is used in further computations. This results in a fuel consumption \eqref{eq:specific} of not more than \SI{34}{\gram/\hour} at \SI{600}{\watt} average power and up to \SI{45.3}{\gram/\hour} at full power.
To fly at least \SI{3}{\hour} at maximum fuel-cell power---to also deliver payload power and be able to climb descend,
hover and recharge hover batteries in-flight---about \SI{140}{\gram} of hydrogen would be desired.

The corresponding Intelligent Energy \ac{TPED} regulator is \SI{0.28}{\kilogram}, 40 by 35mm (diameter x length), 20 to \SI{500}{\bar} `in' and \SI{0.55}{\bar} `out' and is equipped with an electronic shut-off valve, pressure sensors and a standard \SI{8}{\milli\meter} \ac{PCP} fill port.
The fuel-cell system weighs \SI{0.96}{\kilogram} and measures 196 by 100 by \SI{140}{\milli\meter}. It's output voltage ranges from \SI{19.6}{\volt} to \SI{25.2}{\volt}.
It is equipped with a \SI{1800}{\milli\ampere\hour} 6-cell lithium-polymer auxiliary battery of \SI{0.3}{\kilogram} which enables the combined system to deliver
\SI{1400}{\watt} of peak power for a short time.

\subsection{Solid versus Pressure Cylinder}

\ifelspaper
\else

\begin{figure*}[htb]
\begin{subfigure}{.495\textwidth}
  \centering
  \makebox[0.99\textwidth][c]{
  \includegraphics[scale=0.8]{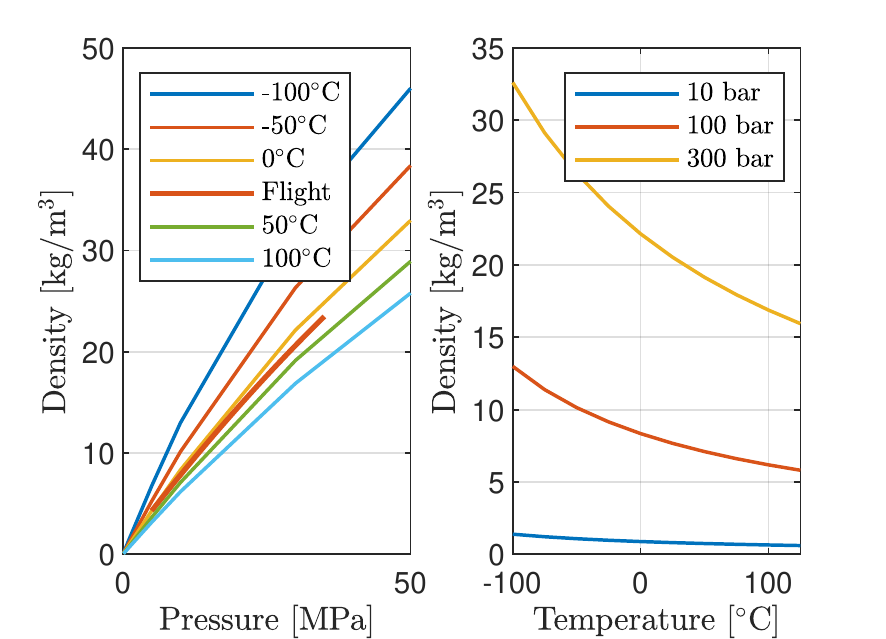}}  
  \caption{Density of pressurized hydrogen in function of temperature \eqref{eq:hydrogen}.}
  \label{fig:hydrodense}
\end{subfigure}
\begin{subfigure}{.495\textwidth}
  \centering
  \makebox[0.99\textwidth][c]{
  \includegraphics[scale=0.8]{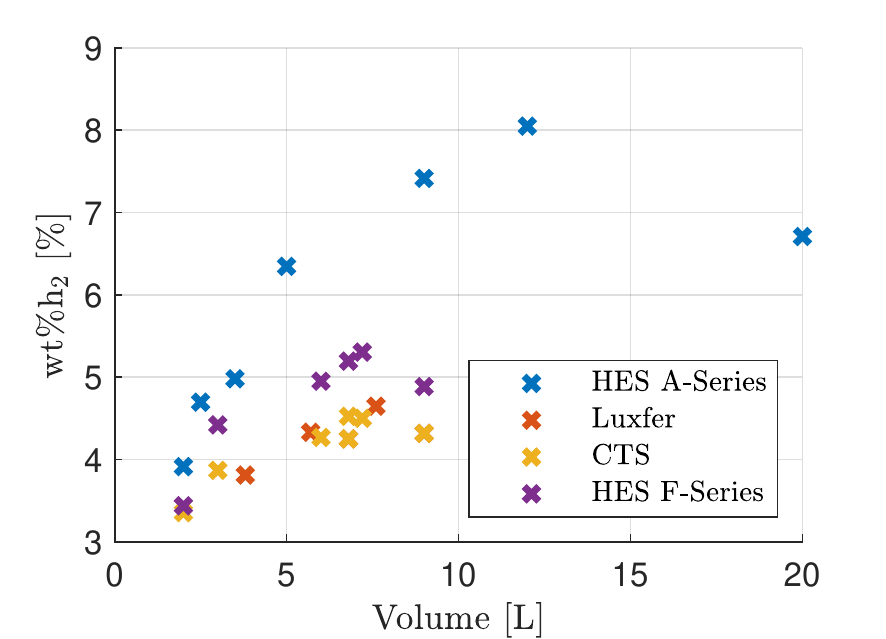} } 
  \caption{Overview of the specific hydrogen weight $wt\%h_2$ for various cylinders that were available at the time of selection.}
  \label{fig:cylinders}
\end{subfigure}
\caption{Hydrogen density and cylinder specific weight. }
\label{fig:cylindershydro}
\end{figure*}

\fi

At room temperature, the main two options to store hydrogen are to store it as a pressurized gas in a pressure cylinder,
or to store it as a chemical solution that releases hydrogen \citep{Zuttel2004_StoreH2}.
Sodium borohydride ($NaBH_4$) has been proposed as hydrogen source to power fuel-cells \citep{Kim2011_soldstate_sodium,Kim2014_solidstate_sodium}.

The downside of pressure cylinders is that they weigh orders of magnitude more than the hydrogen inside them \citep{Lee2020_cylinder_weight_optim}.
But because of sustainability, overal system weight,  off-grid recharge options \citep{Troncoso2014_solarcharge}, price, and availability, the choice was made to use pressure cylinders.

The mass of hydrogen $m_{H_2}$ based on the cylinder volume $V$ and pressure $p$ in flight conditions is fitted as\footnote{\href{https://h2tools.org/hyarc/hydrogen-data/hydrogen-density-different-temperatures-and-pressures}{https://h2tools.org/hyarc/hydrogen-data/hydrogen-density-different-temperatures-and-pressures}}:

\begin{equation}
m_{H_2}=(-0.00002757 p^2 + 0.074969 p + 0.6187)  \cdot V \label{eq:hydrogen}
\end{equation}

It should be noted that the actual value varies with temperature.
At \SI{300}{\bar}, values change from \SI{20.7}{\gram/\liter} at \SI{25}{\degreeCelsius} to \SI{21.2}{\gram/\liter} at \SI{15}{\degreeCelsius} (\fig\ref{fig:hydrodense}) and a \SI{25}{\degreeCelsius} drop in temperature leads to a \SI{7.8}{\percent} increase in hydrogen.
In this paper the more pessimistic values at room temperature are used.
Table~\ref{table:tanks} shows an overview of available cylinder options.
The same data is shown in \fig\ref{fig:cylinders}.
This shows that at time of selection, the lightest cylinders are the HES A-Series and F-Series. 

\begin{figure}[hbt]
\centering
\includegraphics[width=0.80\linewidth]{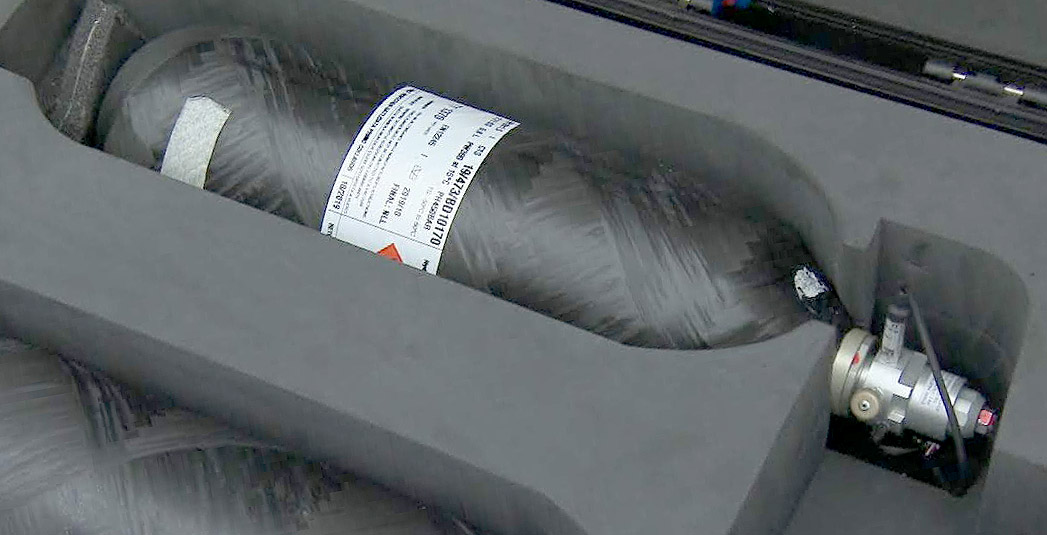}
\caption{\acs{CTS} \SI{6.8}{\liter} \SI{300}{\bar} ultralight Type-4 cylinder in a specially designed certified transport case.} \label{fig:cts_cylinder}
\end{figure}

Unfortunately, due to price, availability and EU certificates, this option was not yet available.
The selected cylinder is the \SI{6.8}{\liter} \ac{CTS} \ac{PET} Liner Type-4 cylinder\footnote{\myurl{http://www.ctscyl.com/prodotti/h2/cts-ultralight-6-8l-300-bar-h2}}.
The graph does however illustrate that doubled weight densities can be expected soon.
A picture of the actual cylinder can be found in \fig\ref{fig:cts_cylinder}.

\subsection{Hydrogen and Safety}

At \SI{300}{\bar}, hydrogen gas weighs about \SI{20.63}{\gram/\liter}.
It is flammable in concentrations from \SI{4}{\percent} up to \SI{75}{\percent} when mixed with air and burns optimally at a concentration of \SI{29}{\percent}.
It has a self-ignition temperature \SI{585}{\degreeCelsius}, but a very low required ignition energy of \SI{17}{\micro\joule}, while human body models show that a person without static protection can easily cause a \SI{40}{\milli\joule} discharge.
To avoid ignition, anti-static shoes and clothes are required when leakages are expected---like for instance during filling.
Refueling should be done at temperatures in between \SI{-20}{\degreeCelsius} and \SI{40}{\degreeCelsius} to stay within tank limitations.
Hydrogen is roughly 14 times lighter than air and therefore easily gets trapped inside rooms.
The area where hydrogen is used should be well ventilated as per \emph{ATEX 153} and for ignition analysis one should refer to \emph{EN1127-1}.
At room temperatures, hydrogen is nearly completely converted to orhto-hydrogen, and no significant heating effect is to be expected when depressurizing it.
When assembling a UAV, applicable regulations include the 
\emph{EU 94/9/EC} (\emph{ATEX 114}) and \emph{ISO 15196} for material properties and their degradation in the presence of hydrogen \cite{Molnarne2019_hydro_risk}. 

\subsection{Cylinder Safety}

A lot of work has already been preformed in the field of pressure tank rupture analysis but most has been done on metal cylinders used for a variety of gasses like \ac{CNG} \citep{Mashayekhi2014_fracture}. 
For composite high pressure cylinders, models and methods have been developed and validated \citep{Bertin2012_type4}, but these do not show all risks of hydrogen cylinder failures.

Actual crush test of composite hydrogen cylinders have been performed by \citep{Mitsuishi2005_crush_type4} to simulate a car crash.
But they used mainly cylinders with aluminum insides (type 3A/B).
The blast of hydrogen cylinders exposed to vehicle fires was also investigated \citep{Molkov2015_blastwave}.
They suggest that the blast from a cylinder failure through fire (with combustion) can throw debris up to \SI{80}{\meter}, but also shows that \SI{35}{\meter} would be a no-harm distance for the shock-wave of a \SI{12}{\liter} \SI{700}{\bar} cylinder, which is much larger than the \ac{UAV} cylinder.

\begin{figure}[hbt]
\begin{subfigure}{.33\linewidth}
  \centering
  \includegraphics[width=.96\linewidth]{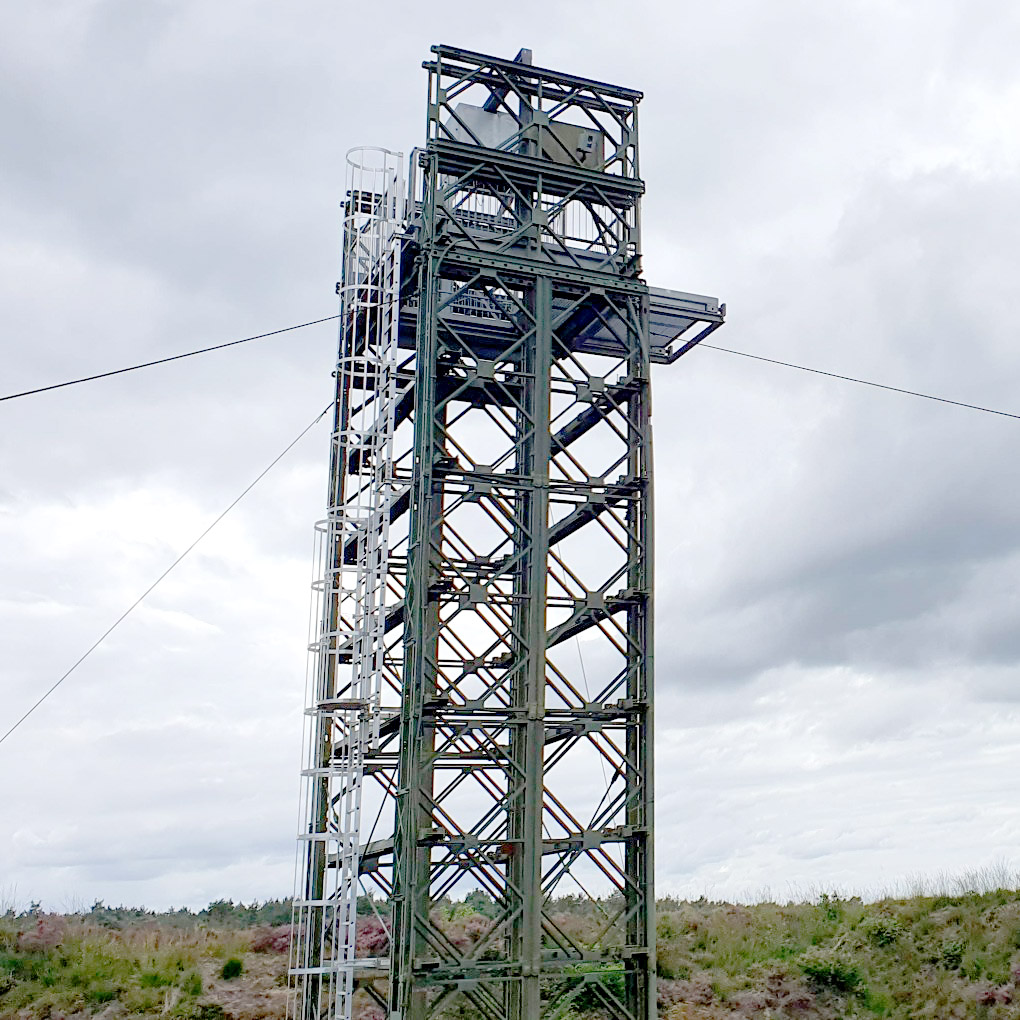}  
  \caption{Drop tower}
  \label{fig:fall:tower}
\end{subfigure}
\begin{subfigure}{.33\linewidth}
  \centering
  \includegraphics[width=.96\linewidth]{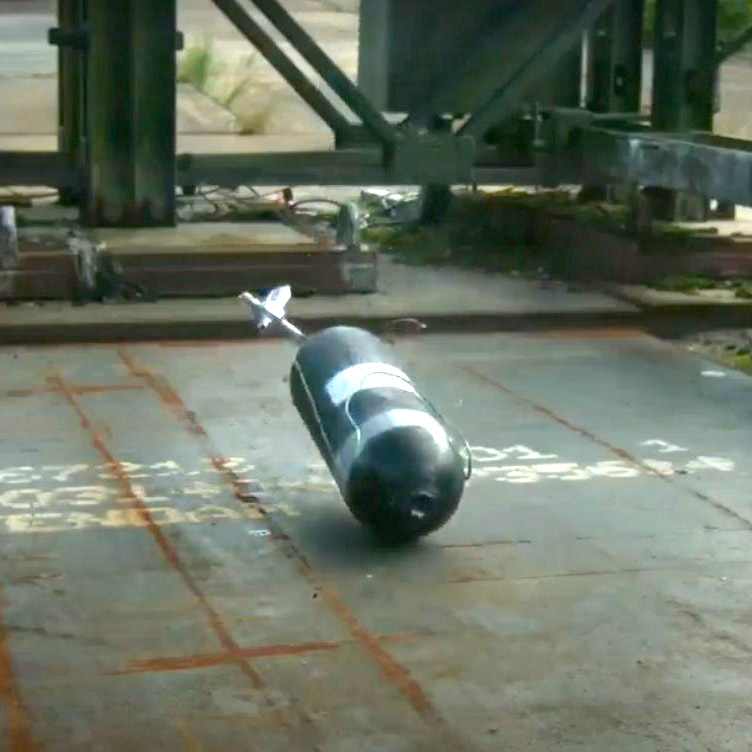}  
  \caption{Tank impact}
  \label{fig:fall:impact}
\end{subfigure}
\begin{subfigure}{.32\linewidth}
  \centering
  \includegraphics[width=.96\linewidth]{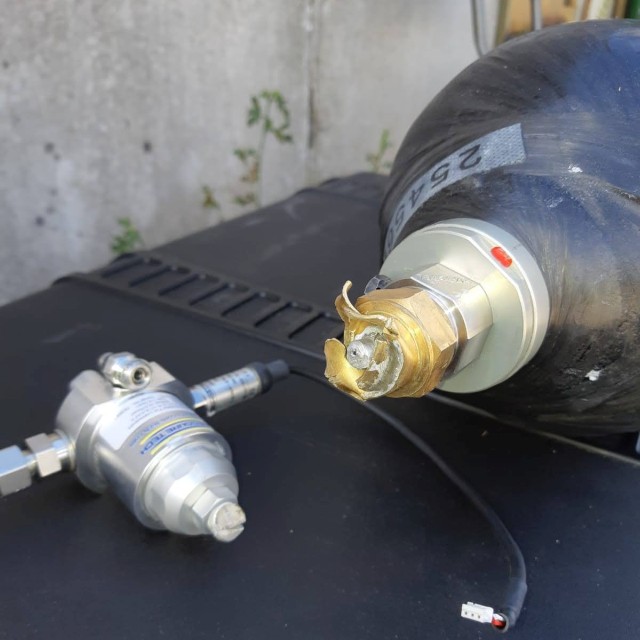}  
  \caption{Damage}
  \label{fig:fall:damage}
\end{subfigure}
\caption{Drop-test of a hydrogen filled cylinder on steel-covered concrete did cause a leak at the regulator but did not visibly damage the cylinder and did not lead to fire nor detonation. }
\label{fig:fall}
\end{figure}

The selected cylinder was tested by the manufacturer according to the \emph{NEN-EN12245+A1}.
Since no data was available about the safety of the combined cylinder and pressure regulator, a drop test was organized that simulated the fall on the metal deck of a ship.
While this does not represent the worst-case scenario of a crash involving hydrogen, it does address the operational scenario in which the hydrogen drone moves away from the ship as soon as possible after take-off and only moves over the ship at low speed and low altitude upon landings.
The test was performed according to the \emph{STANAG 4375}.
The cylinder was dropped from a \SI{12}{\meter} high tower (\fig\ref{fig:fall:tower}) on a metal plate on concrete (\fig\ref{fig:fall:impact}) while filled with \SI{285}{\bar} or about \SI{140}{\gram} of hydrogen.
High-speed camera footage was made and the post impact damage was assessed.
The metal regulator broke (\fig\ref{fig:fall:damage}), which resulted in a leak.
After a few minutes all hydrogen had escaped and the cylinder was inert.

\section{Hybrid lift concept}\label{sec:concept}

Fitting a hydrogen cylinder and fuel-cell in a hybrid UAV poses specific constraints \citep{Furrutter2009_small_fuelcell_100W,Rottmayer2011_fuelcell_naval,Kim2012_hydrogen_quad}.
The large and bulky cylinder highly influences the aerodynamic shape.
To cool the fuel-cell and remove the formed water vapour, sufficient airflow through the fuel-cell radiator is important.
The relatively large weight of the energy supply and payload combined with the weight of the propulsion needed to hover, poses strict limitations on structural weight.
And flying with pressure cylinders and expensive equipment poses stricter redundancy constraints.
This section will go through these challenges.

\subsection{Trade-off}

\begin{figure}[hbt]
\centering
\ifelspaper
\else
\includegraphics[scale=1.0]{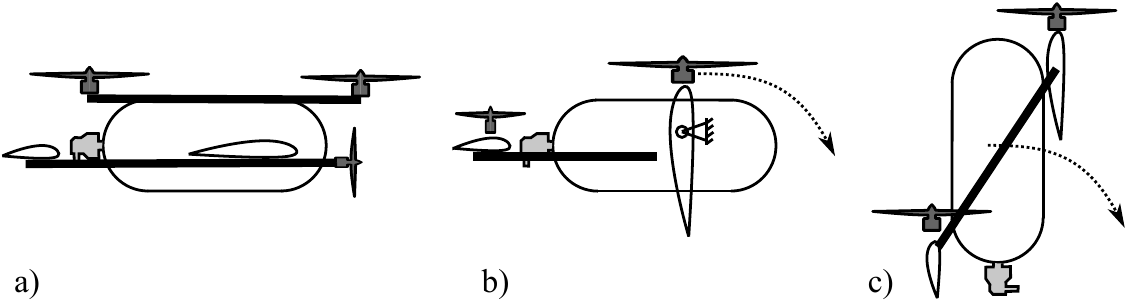}
\fi

\caption{Hybrid lift UAV concepts around a hydrogen cylinder.}\label{fig:concepts:hybrid}
\end{figure}

First a trade off is made between the three main classes of hybrid lift UAVs.
The dual-system \ac{VTOL} \ac{UAV} like quad-planes have a separate propulsion system for hover and forward flight (\fig\ref{fig:concepts:hybrid}a).
To minimize weight, a minimalist hover system is often used as this is dummy weight during the largest part of the flight.
The hover propulsion blows perpendicularly to the wing, and thereby needs additional arms to support the motors at a distance from the wing, which also adds weight and drag in forward flight.
The tilt wing (or tilt-motor) concept (\fig\ref{fig:concepts:hybrid}b) has a mechanism to rotate the entire wing, hereby removing the need for separate motor support arms and re-using part of the propulsion from hover in forward flight.
The downside is increased mechanical complexity, higher mechanical weight and the control complexity of flying with a changing morphology.
A tail-sitter option---shown in \fig\ref{fig:concepts:hybrid}c---re-uses the same motors while keeping mechanical simplicity.
The propulsion can be attached to the wing, which reduces overall structural weight.
As the vast majority of the flight is typically in forward flight, the propulsion can be optimized towards this phase.
The drawbacks are that the UAV must pitch down \ang{90} during the transition and hereby passes through the stall regime of the wing.
Moreover, the cylinder is vertical after landing which makes it prone to tipping over especially on moving platforms like ships.

To minimize structural weight and complexity, while maximally re-using the hover propulsion in forward flight,
the tail-sitter concept was selected for the \nederdrone. 

\subsection{Forward flight drag minimization}

\begin{figure}[hbt]
\centering
\ifelspaper
\else
\includegraphics[scale=1.0]{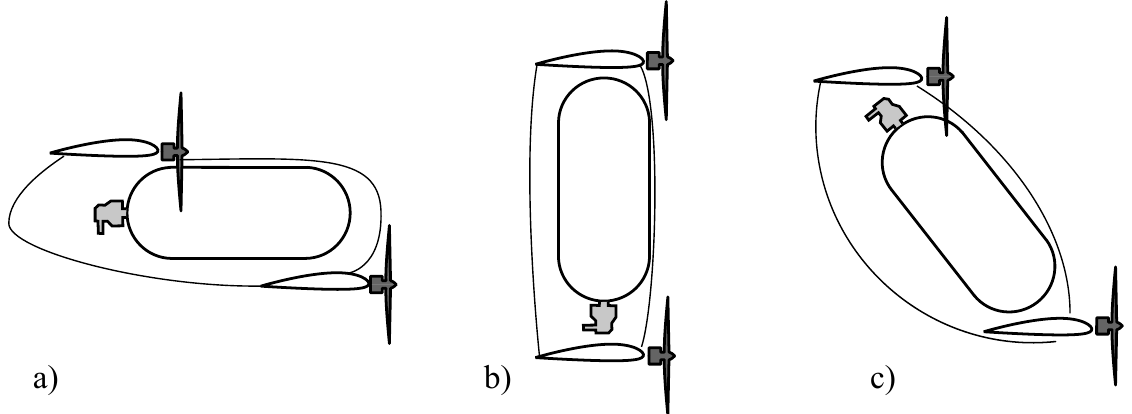}
\fi

\caption{Tail-sitter concepts with several cylinder orientations.}\label{fig:concepts:tailsitters}
\end{figure}

Three variables form a trade-off for the orientation of the cylinder: drag, ground stability and control authority in hover.
The best control authority in hover is achieved by maximizing the distance between the motor center-lines  (\fig\ref{fig:concepts:tailsitters}b).
Hereby, higher control moments can be created through differences in thrust.
After landing the cylinder lies flat and stable on the ground.
But this configuration has the highest fuselage drag in forward flight as the frontal surface is increased.
Moreover, the wings are then placed on top of each other, which cannot result in an aerodynamically stable aircraft in forward flight without S-shaped airfoil or significant wing sweep angle.

Previous work \citep{wagter_2020_nederdrone1} made a compromise and placed the cylinder at a negative \ang{30} angle with the incoming flow (\fig\ref{fig:concepts:tailsitters}c).
The presumed advantages during the landing phase to slowly roll down were found to be insufficient and a landing gear was still needed to protect the propellers.
Moreover the design goal of staying below \SI{600}{\watt} in forward flight could not be achieved.
Therefore the cylinder was placed completely inline with the flow as in  \fig\ref{fig:concepts:tailsitters}a.
This helps to also increase the maximal forward flight speed.

\subsection{Longitudinal stability}

To minimize drag, the cylinder and the fuel-cell are all placed in-line.
This makes the longitudinal distribution of mass in the length of the fuselage significant.
To improve the damping of the short period pitch motion in forward flight,
either a large elevator or a long fuselage is desired \citep{Schmidt1998_flightdynamics}.
This conflicts with the ground stability requirement after landing as a long narrow tail-sitter is at high risk of tipping over and the center of gravity falls from higher in case it does.

The combined requirements are addressed by giving the \nederdrone a short fuselage and a tandem wing configuration.
The tandem wing has the best pitch damping for a given fuselage length.
Moreover, it has a shorter wingspan for a given amount of wing area at a given aspect ratio compared to a conventional large main wing and a small horizontal stabilizer.
These shorter wings help to cope with higher perturbations in hover.

\subsection{Ground stability}

\begin{figure}[hbt]
\centering
\begingroup
  \makeatletter
  \providecommand\rotatebox[2]{#2}
  \ifx\svgwidth\undefined
    \setlength{\unitlength}{184.25196533pt}
  \else
    \setlength{\unitlength}{\svgwidth}
  \fi
  \global\let\svgwidth\undefined
  \makeatother
  \begin{picture}(1,0.63076923)%
    \put(0,0){\includegraphics[width=\unitlength]{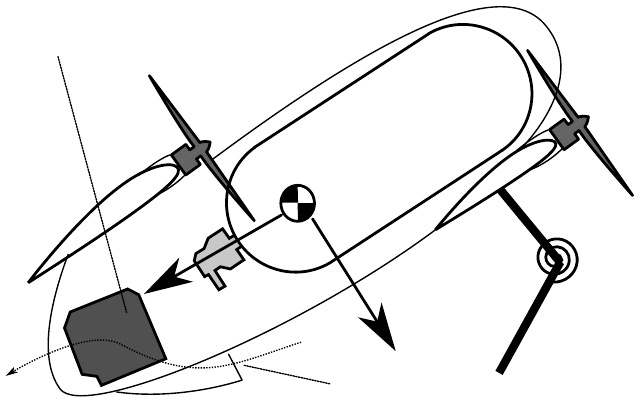}}%
    \put(0.63877981,0.15253902){\color[rgb]{0,0,0}\makebox(0,0)[b]{\smash{X}}}%
    \put(0.27031636,0.24660258){\color[rgb]{0,0,0}\makebox(0,0)[b]{\smash{Z}}}%
    \put(0.62632586,0.41845137){\color[rgb]{0,0,0}\makebox(0,0)[b]{\small\smash{Cylinder}}}%
    \put(0.11478615,0.55852445){\color[rgb]{0,0,0}\makebox(0,0)[b]{\small\smash{Fuel-cell}}}%
    \put(0.52614052,0.0122701){\color[rgb]{0,0,0}\makebox(0,0)[lb]{\small\smash{Cooling}}}%
  \end{picture}%
\endgroup
\caption{The \nederdrone concept: a \emph{drop-down} tail-sitter with an in-flow oriented hydrogen pressurized cylinder, rear-mounted fuel-cell with bottom cooling airflow vent, low front-wing and high tail-wing.}\label{fig:concepts:final}
\end{figure}
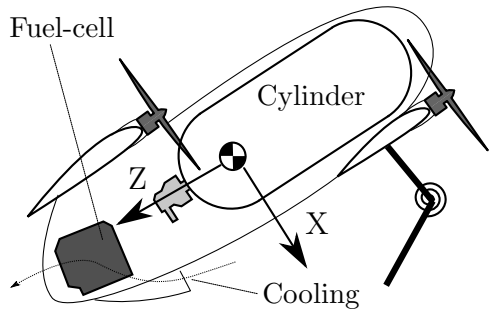

An important property is the stability of the vehicle after landing.
Having a long narrow \ac{UAV} upright containing a high pressure carbon cylinder on a moving platform like a ship after landing is not a stable option.

\begin{figure}[htb]
  \centering
  \includegraphics[width=\linewidth]{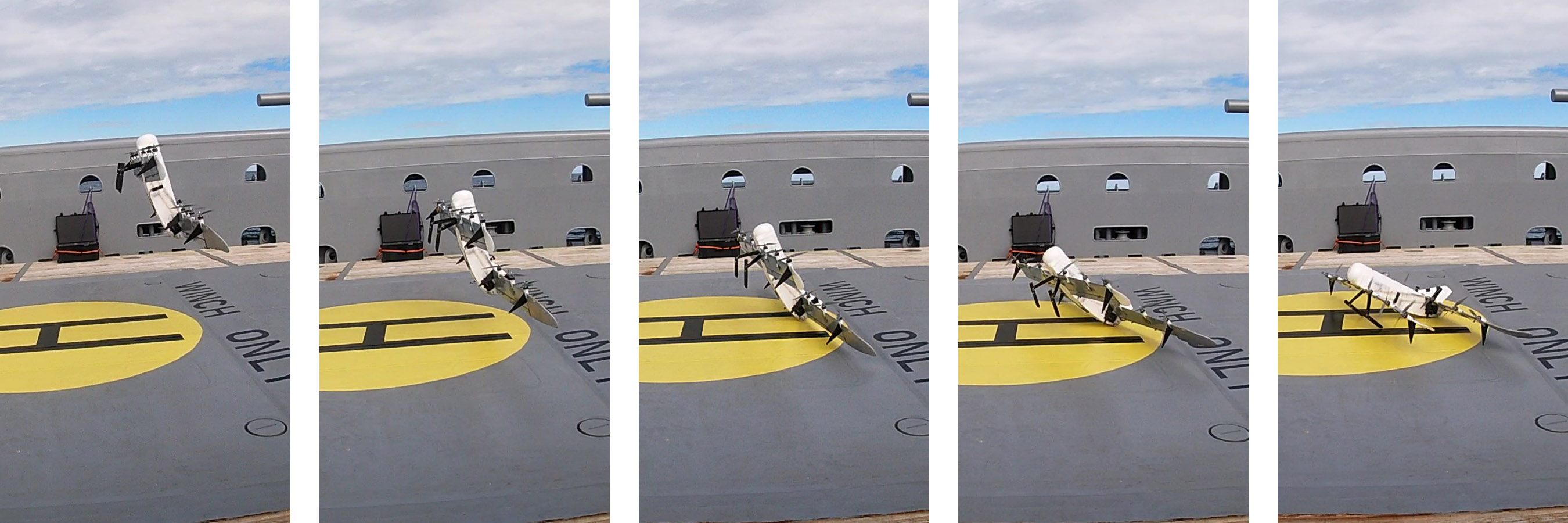}  
\caption{Landing sequence of the \nederdrone on a ship.}\label{fig:landing}
\end{figure}

Inspired by \citep{wagter_2020_nederdrone1} illustrated in \fig\ref{fig:concepts:tailsitters}c, the option was investigated to slowly \emph{drop-down} the nose after landing.
Using the hover propellers, the nose is slowly dropped with the motors on the back wing at minimal thrust.
Once the remaining hover propellers start to point forward beyond a measured angle, the ground friction is overcome and the drone would start to slide forward.
At this point the thrust is cut off and the nose drops down.
To allow this, a sprung landing gear was added which could absorb the last part of the drop.
The result is a \ac{UAV} which lies stably on the ground after landing.

To further minimize the impact of the landing, the center of gravity was moved backwards by choosing a canard configuration with the largest wing at the back.
This places the center of gravity much closer to the ground while in hover.
The resulting landing sequence is shown in \fig\ref{fig:landing}.

\subsection{Take-Off}

\begin{figure}[htbp]
\begin{subfigure}{.49\linewidth}
  \centering
  \includegraphics[width=.98\linewidth]{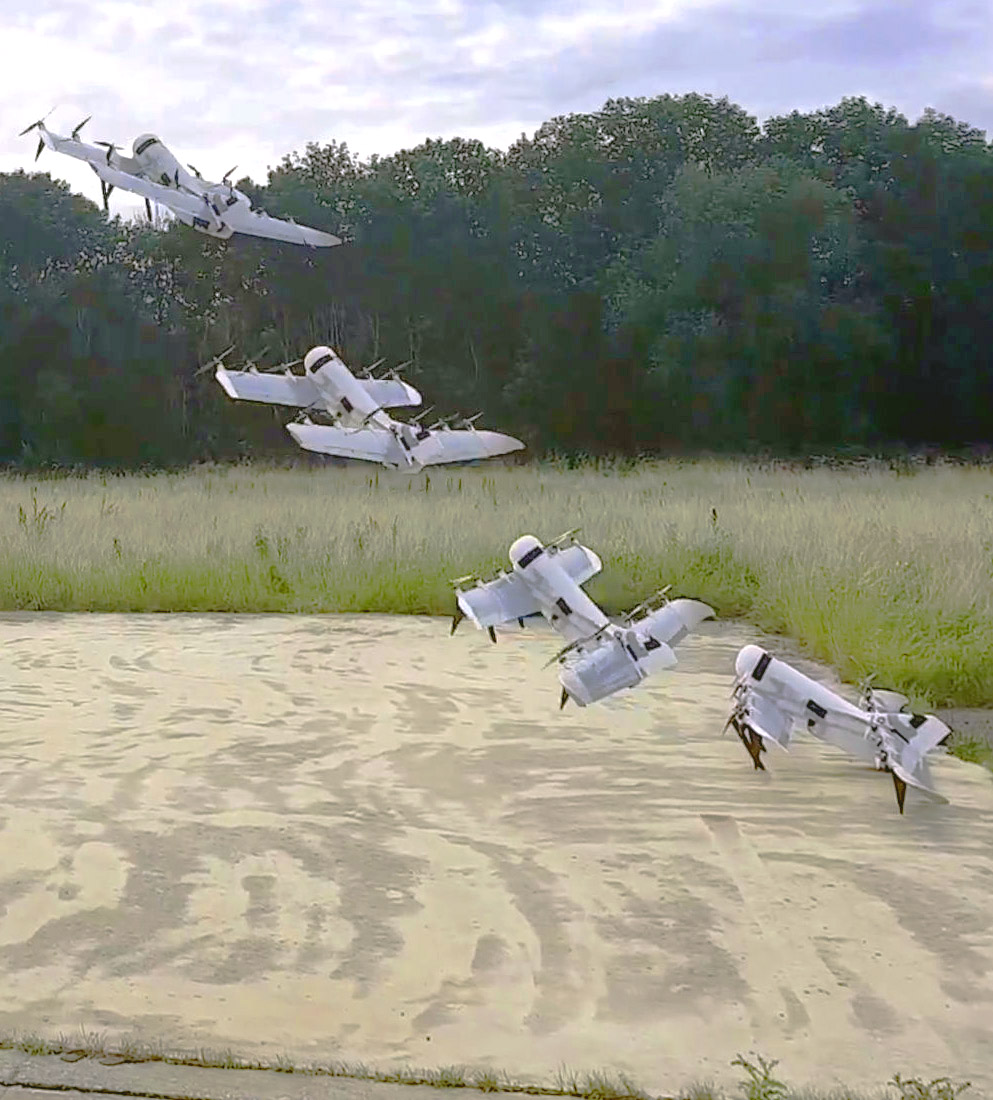}  
  \caption{Take-off with 2m/s wind}
  \label{fig:flight:takeoff1}
\end{subfigure}
\begin{subfigure}{.49\linewidth}
  \centering
  \includegraphics[width=.98\linewidth]{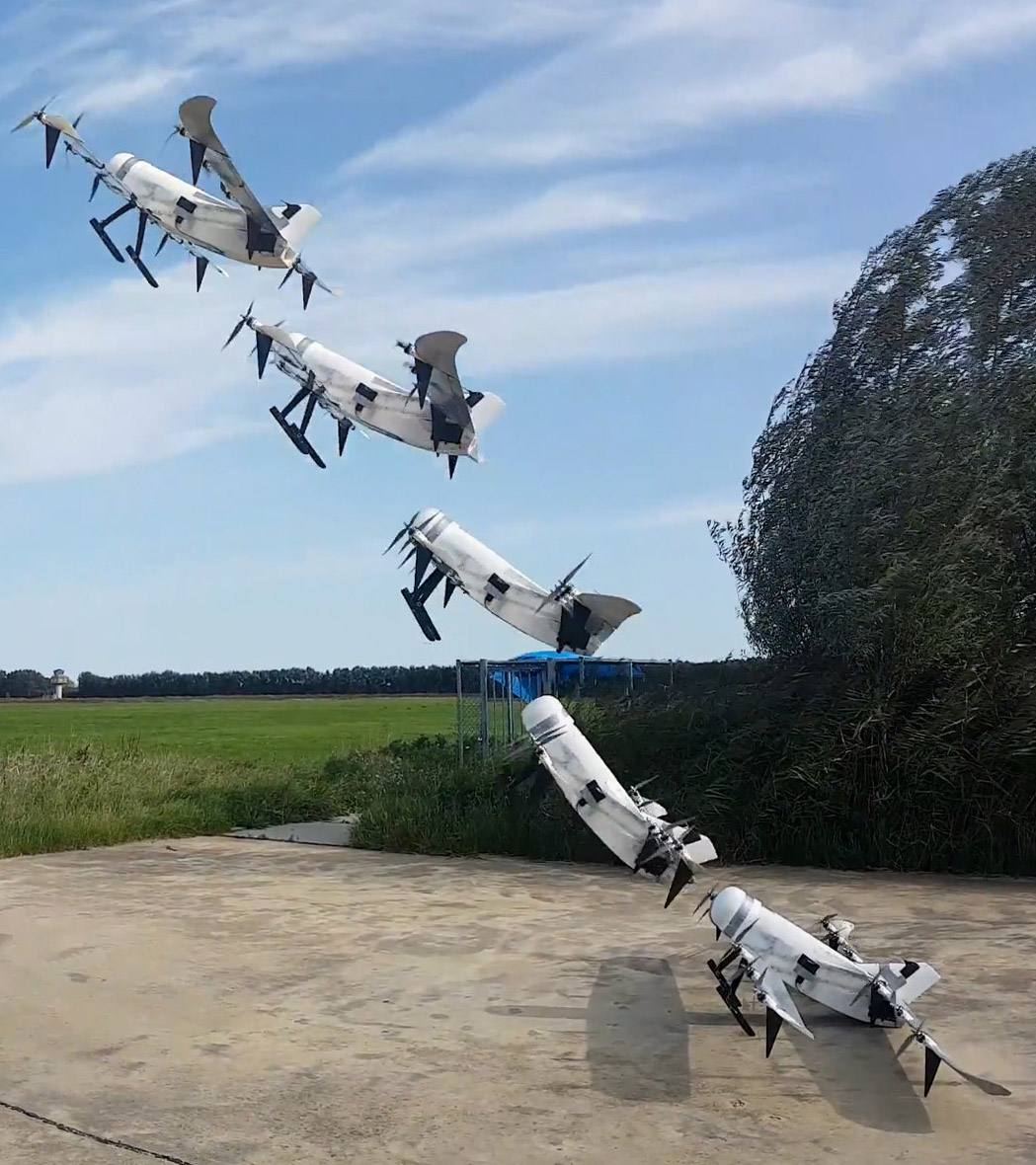}  
  \caption{Take-off with 7m/s wind}
  \label{fig:flight:takeoff2}
\end{subfigure}
\caption{Composite image of the \nederdrone take-off in various wind conditions. }
\label{fig:flight:takeoff}
\end{figure}

The ground stability requirement is conflicting with the vertical take-off requirement of the tail-sitter.
Since the \ac{UAV} is sitting in a \ang{60} nose down from hover, this affects the vertical take-off.
However, test flights showed that even in worst-case `no-wind' conditions the \ac{UAV} only slides less than a foot before taking off as shown in composite image \fig\ref{fig:flight:takeoff}.
The high thrust to weight, the ground effect of the propeller flow over the wing squeezed between the wing and the ground and the spring in the landing gear make the \nederdrone take-off on the spot into what will be referred to as an angled take-off.

\section{Electronics}\label{sec:electronics}

\subsection{Power: concept}

\begin{figure}[hbt]
\centering
\ifelspaper
\else
\includegraphics[scale=1.0]{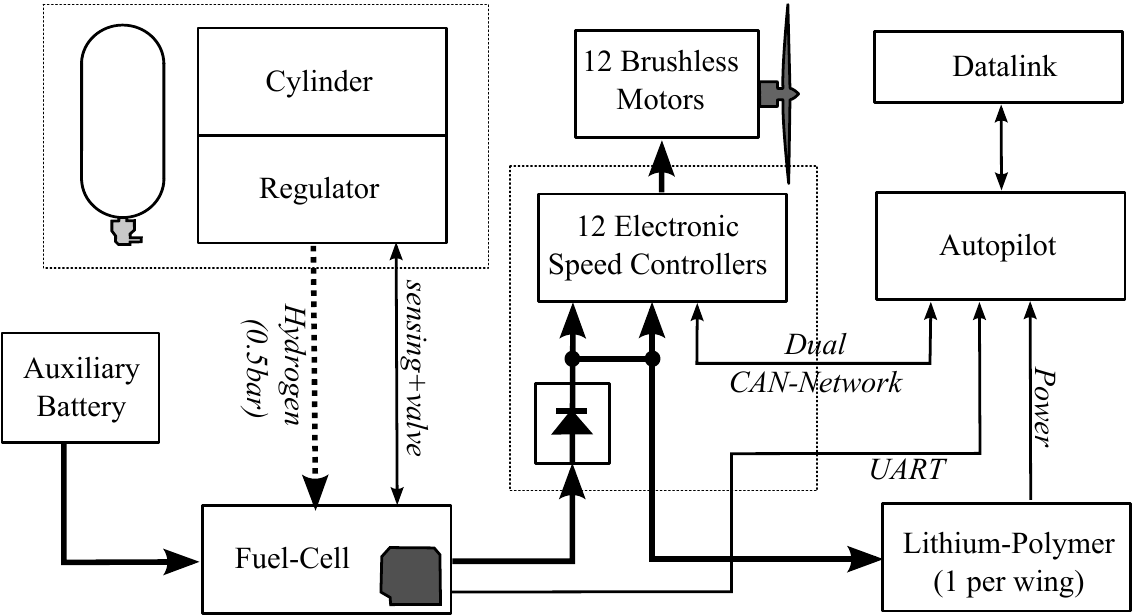}
\fi
\caption{Schematic overview of the main electronics and wiring.}\label{fig:elect}
\end{figure}

Hovering a \SI{10}{\kilogram} platform in gusts while the propulsion is optimized for the much longer forward flight phase is taking more than the \SI{1400}{\watt} maximum of the fuel-cell system.
To complement the fuel-cell during the short high power phases, high C-rating lithium polymer batteries are added to the \nederdrone.
The fuel-cell provides a nominal output voltage of \SI{25}{\volt}, which drops when the load increases.
Six cell lithium-polymer batteries recommendation state that they can safely be charged up to \SI{25.2}{\volt}, which matches perfectly with the fuel-cell voltage range including a safety margin of \SI{0.2}{\volt} to prevent over-charge.
To minimize weight, the batteries are connected directly to the motors in parallel to the fuel-cell.
But to prevent that the hover batteries would feed current into the fuel-cell, the fuel-cell current is run through a set of small diodes at each motor.
This allows the very high currents to go form the lithium battery directly to the \ac{ESC} without loss and allows the fuel-cell to re-charge the batteries to no more than \SI{25}{\volt} minus the diode forward drop voltage of about \SI{0.2}{\volt}.
This means the lithium-polymer batteries are charged up to about \SI{24.8}{\volt} which corresponds to at least \SI{95}{\percent} full.
The power and control wiring is depicted in \fig\ref{fig:elect}.

The selected hover battery consists of four \emph{Extron X2} \SI{4500}{\milli\ampere\hour} 6S 1P lithium-polymer batteries with a nominal voltage of \SI{22.2}{\volt} and a discharge rate of 25C to 50C.
They contain just under \SI{100}{\watt\hour} of additional energy at \SI{640}{\gram} each and can supply continuous \SI{90}{\ampere} and \SI{180}{\ampere} burst currents.
The four batteries are placed as near as possible to the four wings to supply the 3 motors on each with short high power wires.
These 4 extra batteries combined provide sufficient current for the most demanding conditions and provide enough energy to fly for at least \SI{20}{\minute} in case the fuel-cell fails in-flight.

\subsection{Control bus: Aerospace CAN}

For redundancy and structural weight distribution purposes the \nederdrone has twelve motors.
To reduce wiring and connector failures and create a system which is still able to fly even if any of the wires would fail, 
the power and control wires are duplicated.
This would lead to 24 control wires going to the 12 motors, excluding the motor status feedback wires and dual power bus.
To reduce this large amount of wiring and weight, a control network is used (See \fig\ref{fig:elect}).

The \acf{CAN} is an automotive industry technology that has been proposed as a low cost solution in several aerospace applications \citep{SPARKS1997_CAN_lowcost}.
The increasingly popular \citep{Ravi2019_uavcan} \emph{UAVCAN}\footnote{\myurl{https://uavcan.org/}} implementation was selected with custom messages.
The resulting system is a setup where any control or power wire can be cut without dramatic consequences, while the weight and complexity is kept to a minimum.



\section{Control}\label{sec:control}

In selecting a controller for the \nederdrone, its special challenges were considered.
First, though the \nederdrone does not `sit on its tail' when landed, in terms of flight mechanics it does behave like a tail-sitter.
The control of tail-sitters is complicated by the wings that operate at large angles of attack during slow flight.
This makes it difficult to model the lift, drag and pitching moment accurately.
It can also lead to sudden changes in the aerodynamic forces and moments when the flow over the wings suddenly stalls or re-attaches.
Tail-sitters are typically susceptible to wind gusts, due the large exposed surface during hovering flight.
These disturbances need to be compensated by the controller.
Second, the design of the \nederdrone is somewhat unorthodox with its tandem-wing configuration, and the slipstream from the front wing can hit the back wing.
This is expected to produce a complex interaction at certain angles of attack (consider \fig\ref{fig:concepts:final} and imagine a horizontal velocity to the right), which would be hard to predict.
Third, the experimental nature of the project required a control method that could be easily adapted to changes made to the platform, without needing new wind tunnel tests.

To cope with these challenges, \ac{INDI} was proposed as the method of control, because of its successful implementation on the Cyclone tail-sitter \ac{UAV}, which had similar challenges \citep{Smeur2019_cyclone_indi}.

\subsection{Cascaded INDI Control}

\ac{INDI} is a control method that makes use of feedback of linear and angular acceleration to replace much of the modeling needs, since these signals provide direct information on the forces and moments that act on the vehicle \citep{Smeur2016_indi}.
The angular acceleration can be obtained through differentiation of the gyroscope signal, and the linear acceleration is directly measured with the accelerometer.
Based on the difference between desired and measured linear and angular acceleration, control \emph{increments} are calculated using the control effectiveness.
Because disturbances are directly measured with the accelerometer and the gyroscope, they can be counteracted very effectively.
The disturbance rejection properties of \ac{INDI} have been shown theoretically and experimentally in previous research \citep{smeur2018_cascaded_indi}.

\begin{figure*}[hbt]
\makebox[0.95\textwidth][c]{
\ifelspaper
\else
\def\svgwidth{390pt}
\fi
\begingroup
  \makeatletter
  \providecommand\rotatebox[2]{#2}
  \ifx\svgwidth\undefined
    \setlength{\unitlength}{408.18896484pt}
  \else
    \setlength{\unitlength}{\svgwidth}
  \fi
  \global\let\svgwidth\undefined
  \makeatother
  \begin{picture}(1,0.20138891)%
    \put(0,0){\includegraphics[width=\unitlength]{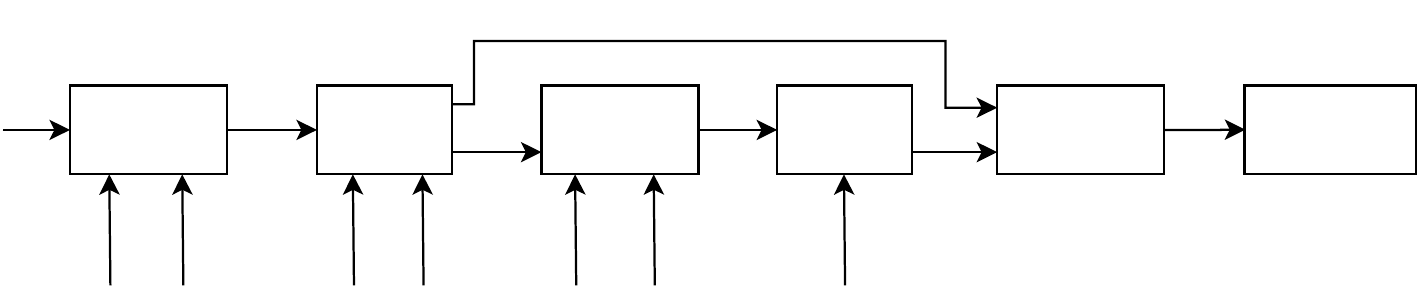}}%
    \put(0.1911424,0.13559283){\color[rgb]{0,0,0}\makebox(0,0)[b]{\smash{\small$\ddot{\xi}_r$}}}%
    \put(0.34954321,0.10423481){\color[rgb]{0,0,0}\makebox(0,0)[b]{\smash{\small$\eta_r$}}}%
    \put(0.5237841,0.13559283){\color[rgb]{0,0,0}\makebox(0,0)[b]{\smash{$\dot{\omega}_r$}}}%
    \put(0.67109685,0.03524715){\color[rgb]{0,0,0}\makebox(0,0)[b]{\smash{\scriptsize$\begin{pmatrix} L \\ M \\ N \end{pmatrix}$}}}%
    \put(0.503192,0.18262987){\color[rgb]{0,0,0}\makebox(0,0)[b]{\smash{\small$T$}}}%
    \put(0.01667325,0.13500816){\color[rgb]{0,0,0}\makebox(0,0)[b]{\smash{\small$\xi_r$}}}%
    \put(0.85010766,0.11999721){\color[rgb]{0,0,0}\makebox(0,0)[b]{\smash{\small$u$}}}%
    \put(0.06600577,0.01956814){\color[rgb]{0,0,0}\makebox(0,0)[b]{\smash{\small$\xi$}}}%
    \put(0.11352601,0.01956814){\color[rgb]{0,0,0}\makebox(0,0)[b]{\smash{\small$\dot{\xi}$}}}%
    \put(0.23391062,0.01956814){\color[rgb]{0,0,0}\makebox(0,0)[b]{\smash{\small$\ddot{\xi}_f$}}}%
    \put(0.39389544,0.01956814){\color[rgb]{0,0,0}\makebox(0,0)[b]{\smash{\small$\eta$}}}%
    \put(0.4445837,0.02113604){\color[rgb]{0,0,0}\makebox(0,0)[b]{\smash{\small$\omega$}}}%
    \put(0.58080839,0.02113604){\color[rgb]{0,0,0}\makebox(0,0)[b]{\smash{\small$\dot{\omega}_f$}}}%
    \put(0.27509483,0.01956814){\color[rgb]{0,0,0}\makebox(0,0)[b]{\smash{\small$\eta_f$}}}%
    \put(0.94053518,0.10601661){\color[rgb]{0,0,0}\makebox(0,0)[b]{\smash{\scriptsize\nederdrone}}}%
    \put(0.0491263,0.12852022){\color[rgb]{0,0,0}\makebox(0,0)[lt]{\begin{minipage}{0.1119878\unitlength}\centering \scriptsize Linear controller\end{minipage}}}%
    \put(0.38105049,0.1279908){\color[rgb]{0,0,0}\makebox(0,0)[lt]{\begin{minipage}{0.11199361\unitlength}\centering \scriptsize Linear controller\end{minipage}}}%
    \put(0.22160198,0.12852021){\color[rgb]{0,0,0}\makebox(0,0)[lt]{\begin{minipage}{0.09721494\unitlength}\centering \scriptsize INDI\\outerloop\end{minipage}}}%
    \put(0.54812328,0.12852022){\color[rgb]{0,0,0}\makebox(0,0)[lt]{\begin{minipage}{0.09585546\unitlength}\centering \scriptsize INDI\\innerloop\end{minipage}}}%
    \put(0.70341463,0.12946361){\color[rgb]{0,0,0}\makebox(0,0)[lt]{\begin{minipage}{0.1170943\unitlength}\centering \scriptsize Control\\allocation\end{minipage}}}%
  \end{picture}%
\endgroup
}
\caption{A schematic overview of the cascaded INDI control approach of the \nederdrone.}\label{fig:control_diagram}
\end{figure*}
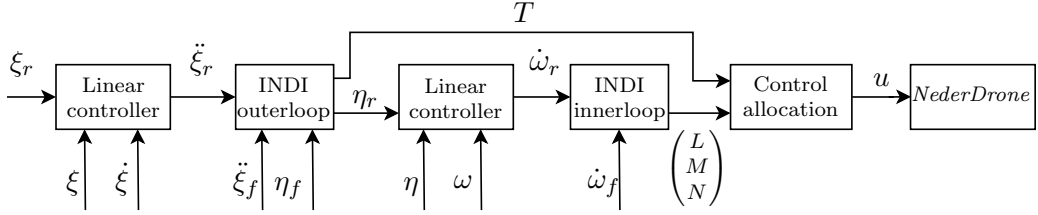

The general structure of the controller is given in \fig\ref{fig:control_diagram}, in which $\xi$ is the position, $\eta$ is the attitude, and $\omega$ the angular rate of the vehicle.
The control moments are denoted by $L$, $M$ and $N$, the total thrust is $T$, and the commands to the servos and motors is $u$.
Signals that are filtered with a low pass filter have a subscript $f$.
From this figure the cascaded structure becomes clear, with an inner and an outer \ac{INDI} loop.

\subsection{Structural Modes}

\begin{figure}[hbt]
\centering
\makebox[0.95\linewidth][c]{\includegraphics[scale=0.8]{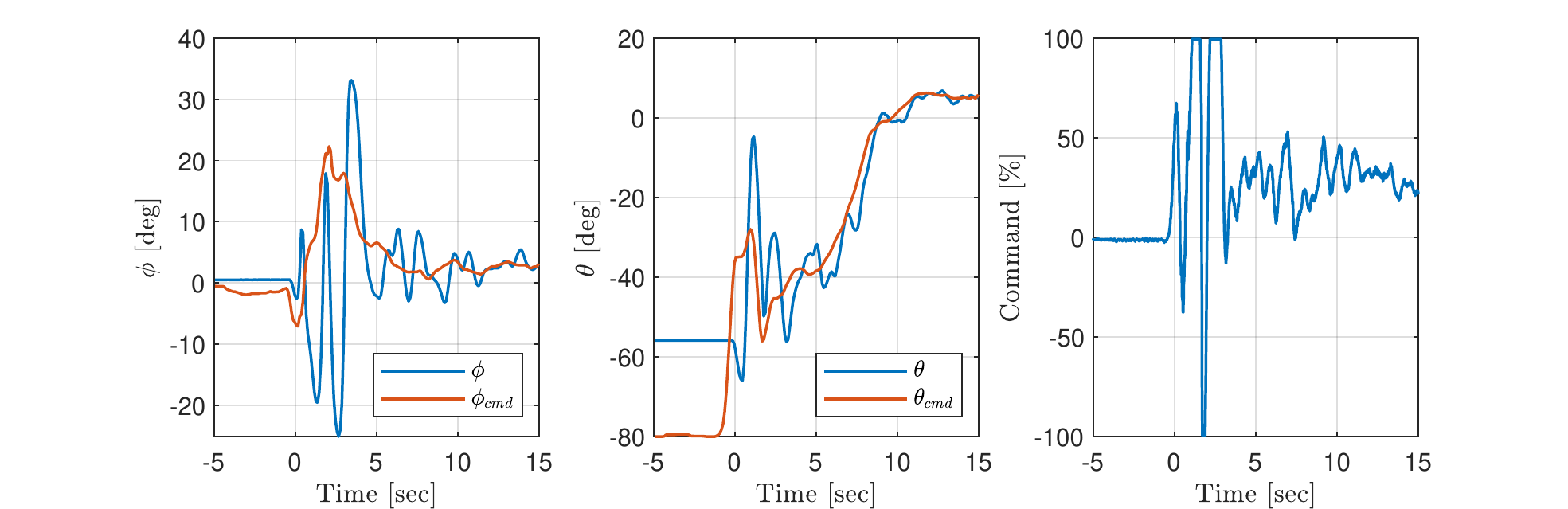}}
\caption{Time sequence of a take-off ($t$=0) with a tip propeller spinning in reverse, causing a very large roll disturbance.
The \ac{INDI} controller needs \SI{100}{\percent} deflections but finds the required trim command within seconds.
Notice the \ang{55} nose down $\theta$ when standing on the ground.
}
\label{fig:control:reverseprop}
\end{figure}

During the test flights, it was found that the frequency of some of the structural modes---in particular the longitudinal torsion mode---was relatively low.
To avoid interaction between the controller and the structural modes, the common procedure in aerospace is to make sure that the controller has a sufficiently small open loop gain at the structural resonance frequency \cite{Preumont2018_ctrl_vibration}.
This can be achieved by including a low pass or a notch filter on the relevant feedback signals.
For the \ac{INDI} inner loop, there is already a low pass filter, since the angular acceleration signal is typically noisy due to high frequency vibrations coming from the motors.
Though including a separate notch filter to dampen the structural mode could result in an overall lower phase lag, this also requires knowledge of the frequency of the structural mode.
To keep the design simple, the cutoff frequency of the second order Butterworth low pass filter that was already in place was set to \SI{1.5}{\hertz} for the pitch rate and yaw rate, and to \SI{0.5}{\hertz} for the yaw rate and linear acceleration.
It should be noted that these filter cutoff frequencies are relatively low, and lead to reduced disturbance rejection performance.

Nevertheless, to illustrate the ability of \ac{INDI} to handle large disturbances, \fig\ref{fig:control_diagram} shows a \nederdrone take-off with one tip propeller turning in reverse direction---thus pushing backwards instead of forwards.
The INDI stabilizes the \nederdrone within seconds although requiring a \SI{25}{\percent} roll command to compensate.


\section{Aerodynamics}\label{sec:windtunnel}

\begin{figure}[hbt]
\centering
\includegraphics[width=0.9\linewidth]{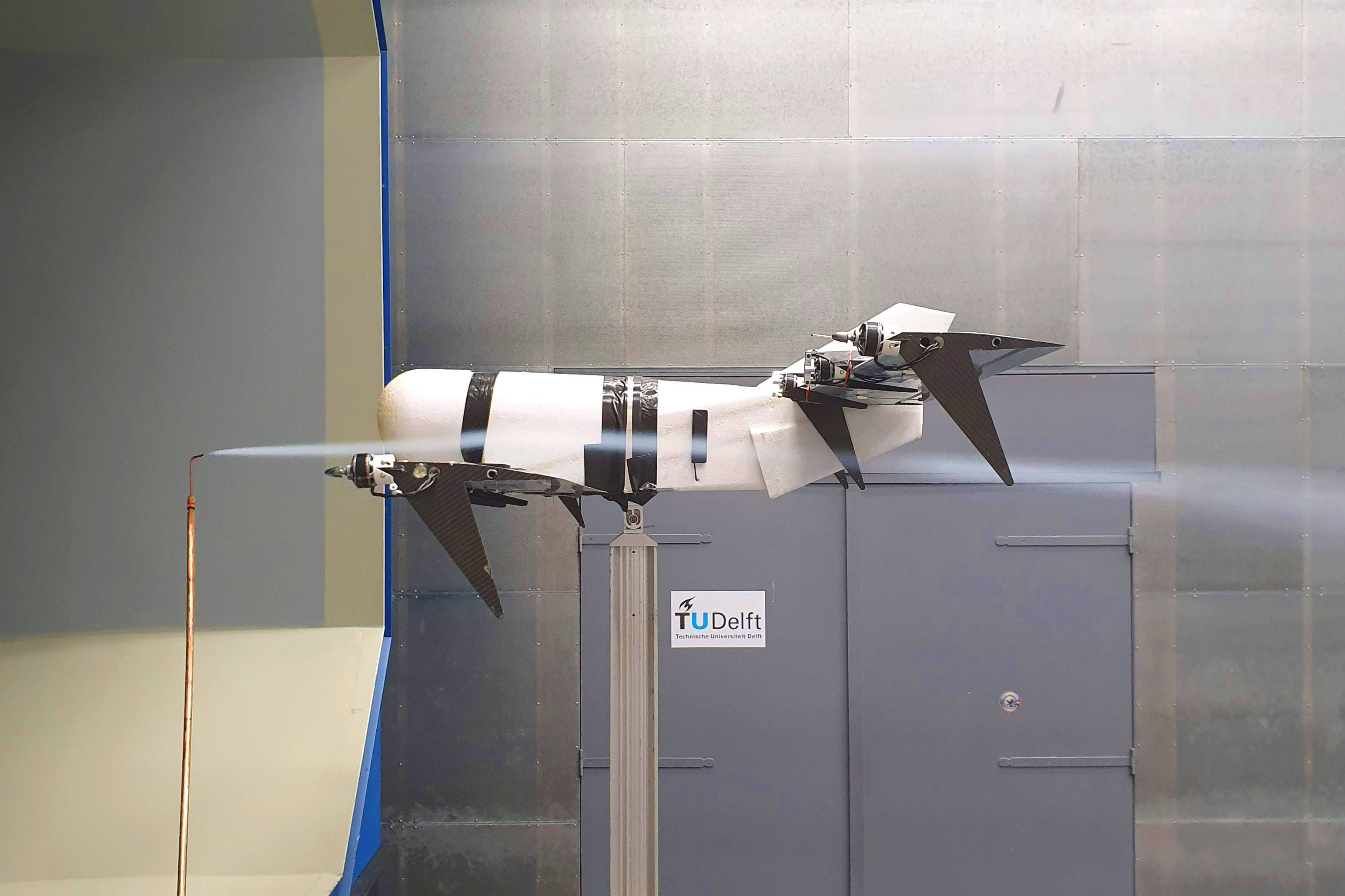}
\caption{
\nederdrone placed in the TUDelft open-jet wind-tunnel.
Smoke analysis shows that in forward flight no interference exists between the wings and that cooling air reaches the rear inlet to cool the fuel-cell.}\label{fig:windtunnel}
\end{figure}

\begin{figure}[hbt]
\centering
\includegraphics[scale=0.8]{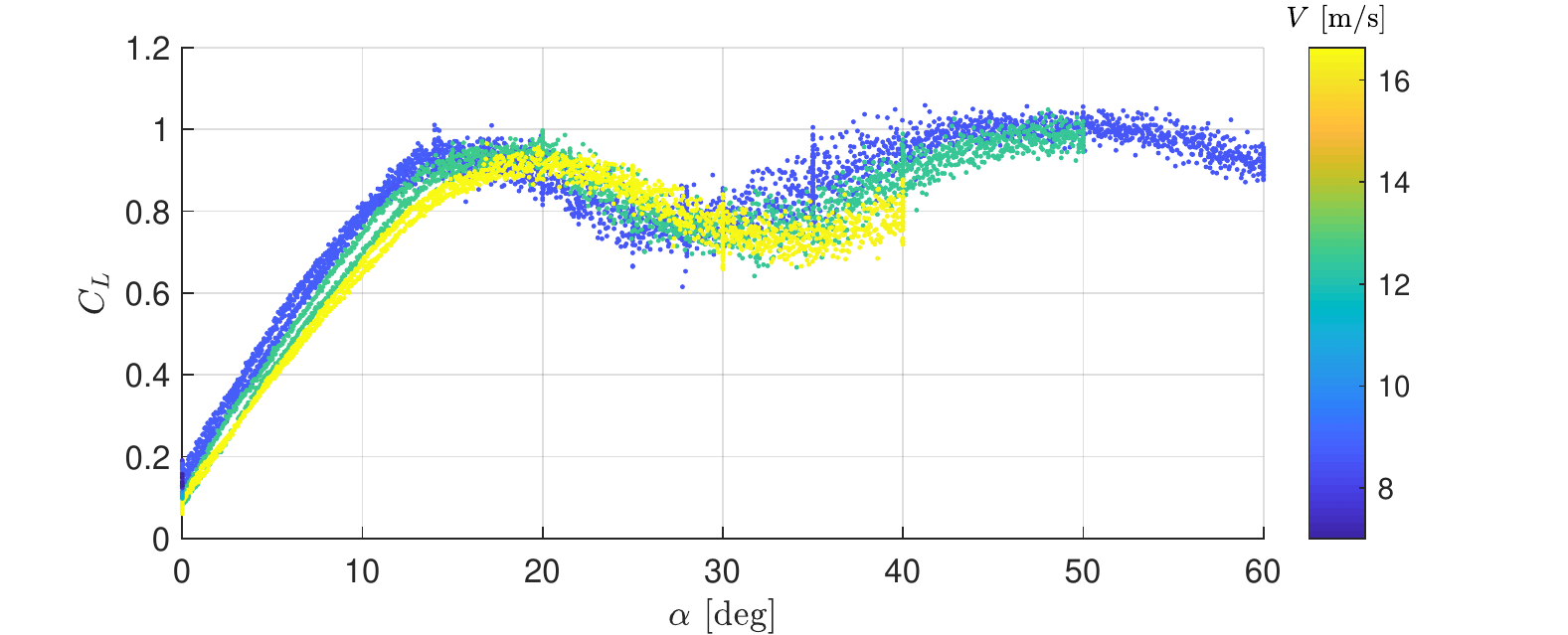}
\caption{Lift curve (lift coefficient $C_L$ in function of angle of attack $\alpha$) of the \nederdrone as measured during wind tunnel testing. The stall starts at about \ang{15} but is very gentle. This is important during the transition phase of the tail-sitter \ac{UAV} as abruptly changing lift forces complicate the control.}\label{fig:liftcurve}
\end{figure}

The airfoil for the wings has been chosen for a good compromise between `gentle stall' and `low drag throughout the lift curve' using the \emph{XFoil} module within XFLR5\footnote{\myurl{http://www.xflr5.tech/}}.
It is based on a \emph{MH32} airfoil, but was modified to allow construction from \ac{EPP}.
Wind-tunnel measurements were performed in the TUDelft \ac{OJF} \cite{Vermeer2001_windtunnel}, in which the full scale \nederdrone can be tested (See \fig\ref{fig:windtunnel}).
The gentle stall was confirmed and is shown in \fig\ref{fig:liftcurve}.

\section{Results}\label{sec:flight}

\begin{figure}[hbt]
\centering
\makebox[0.48\textwidth][c]{\includegraphics[scale=0.8]{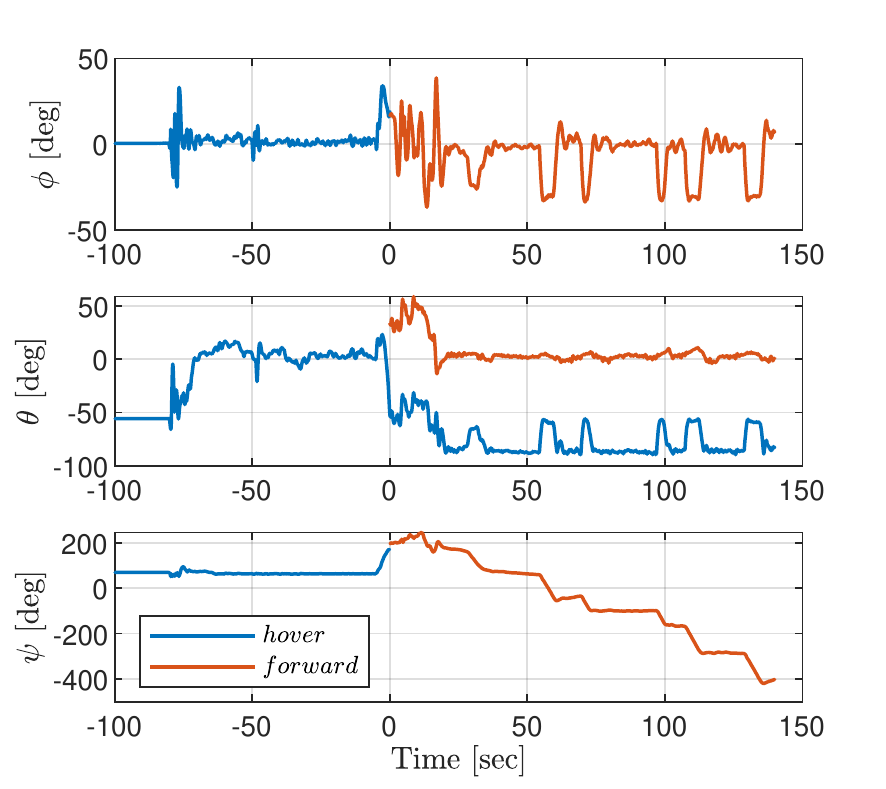}}
\makebox[0.48\textwidth][c]{\includegraphics[scale=0.8]{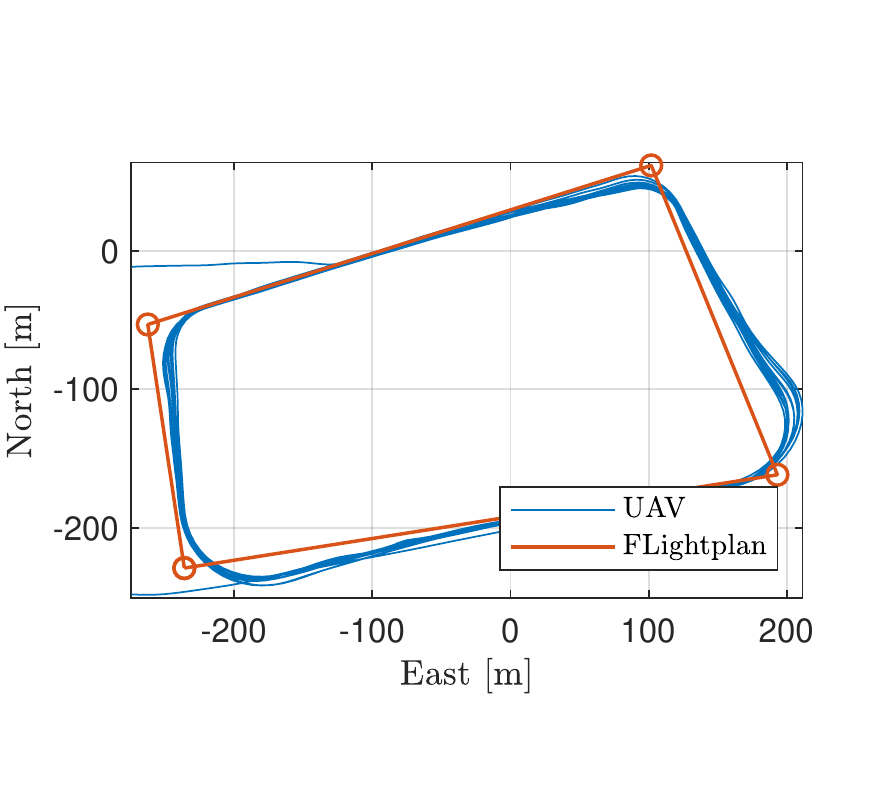}}
\caption{Test flight with take-off, hover, transition to forward and forward flight through a set of waypoints. The Euler angles in the plots swap from the hover frame to the forward frame upon transition. Onboard computations are performed in quaternions.} \label{fig:control:flightpath}
\end{figure}

The concept was built and tested in real flight.
The wings are made of \ac{EPP} cut with hot-wire and strengthened with dual carbon spars.
Carbon ribs connect the spars to the motor mounts.
Various parts like the motor mounts were built with the increasingly popular and powerful 3D printing technology \citep{Goh2017_additive_manufact}.
To withstand the motor heat, the motor mounts were printed from high temperature resistant Ultimaker CPE+ filament.
The autopilot software is the Paparazzi-UAV autopilot \citep{Brisset2006_paparazzi_solution,Gati2013_pprz} project, which has support for various key features like low-level \ac{CAN} drivers to \ac{INDI} control implementations for hybrid aircraft, together with the ability to easily create custom modules to interface with the fuel-cell systems.
The autopilot hardware is the Pixhawk PX4 \emph{MBS-ENTB-24} board.
The used motors are 12 T-Motor \emph{MN3510-25-360} motors equipped with \emph{APC 13x10} propellers.
Servos are the waterproof \emph{HS-5086WP} metal gear, micro digital waterproof servos.
Telemetry is exchanged via the \emph{HereLink} system.

\subsection{Battery-only flight testing}

Before flying with hydrogen, a mock-up hydrogen system was 3D printed and filled with batteries and metal to achieve the exact component weight.
The mock-up cylinder was equipped with a \SI{21}{\ampere\hour} \SI{1.865}{\kilogram}
6S6P lithium-ion battery (\emph{NCR18650GA}) to simulate the power delivered from the fuel-cell.
This allowed to safely fly over \SI{30}{\minute}.
A sample flight is shown in \fig\ref{fig:control:flightpath} in which the pitch angle on the ground, take-off, hover, transition, forward flight and forward turns can be seen.
Once the \nederdrone with mock-up hydrogen components had flown dozens of flight hours successfully including many test flight from a ship, it was equipped with the hydrogen systems.

\subsection{The first ever hybrid lift hybrid energy hydrogen flight at sea}

\begin{figure}[hbt]
\centering
\includegraphics[width=0.90\linewidth]{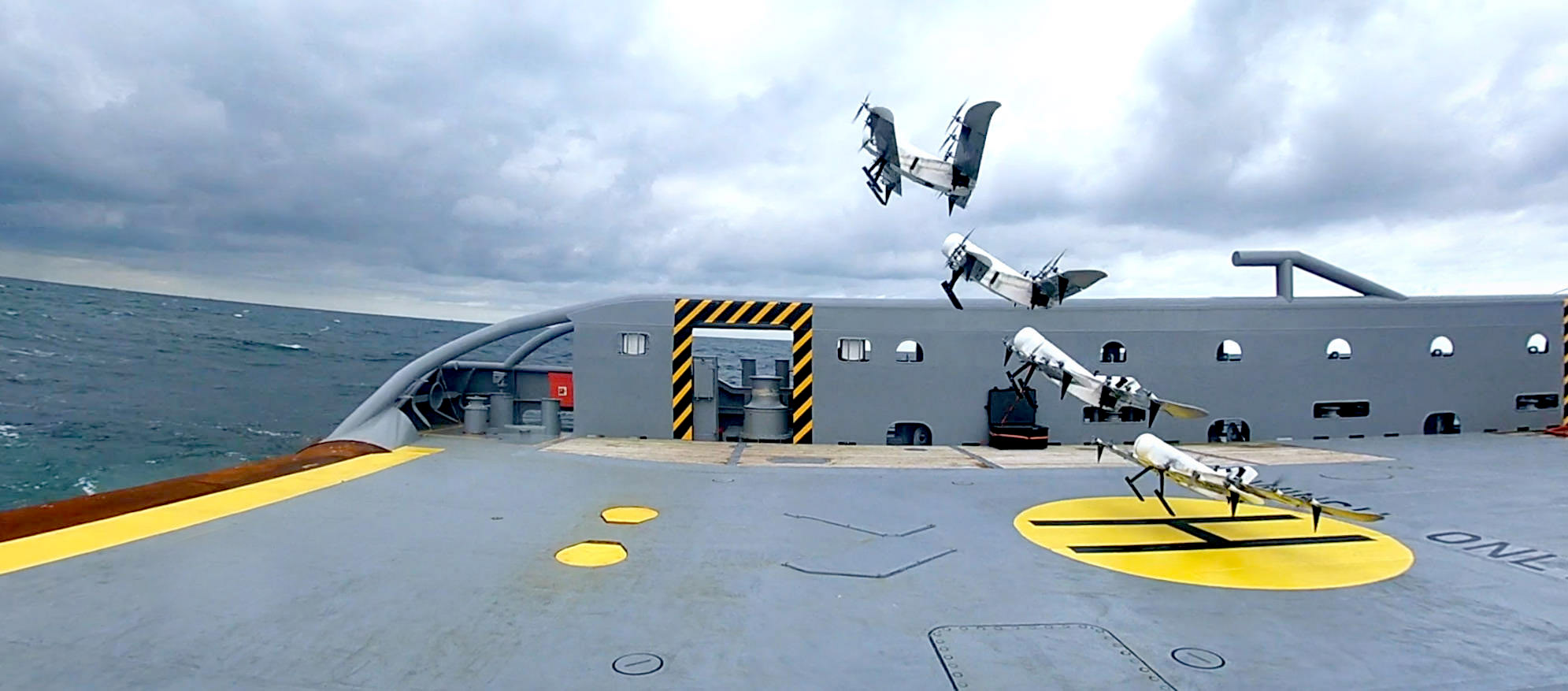}
\caption{Composite image from take-off at sea from the moving coast-guard vessel the \emph{Guardian}.}\label{fig:flight:takeof}
\end{figure}

To demonstrate the capabilities of the \nederdrone we performed a test flight at sea in real world conditions.
On September 30$^{th}$ 2020, the \nederdrone with fuel cell took off from a sailing coast guard ship in moderate wind conditions with 20 knots of wind. 
The flight lasted 3 hour and 38 minutes.
A composite image of the take-off is shown in \fig\ref{fig:flight:takeof}.

\begin{figure}[hbtp]
\centering
\includegraphics[width=0.90\linewidth]{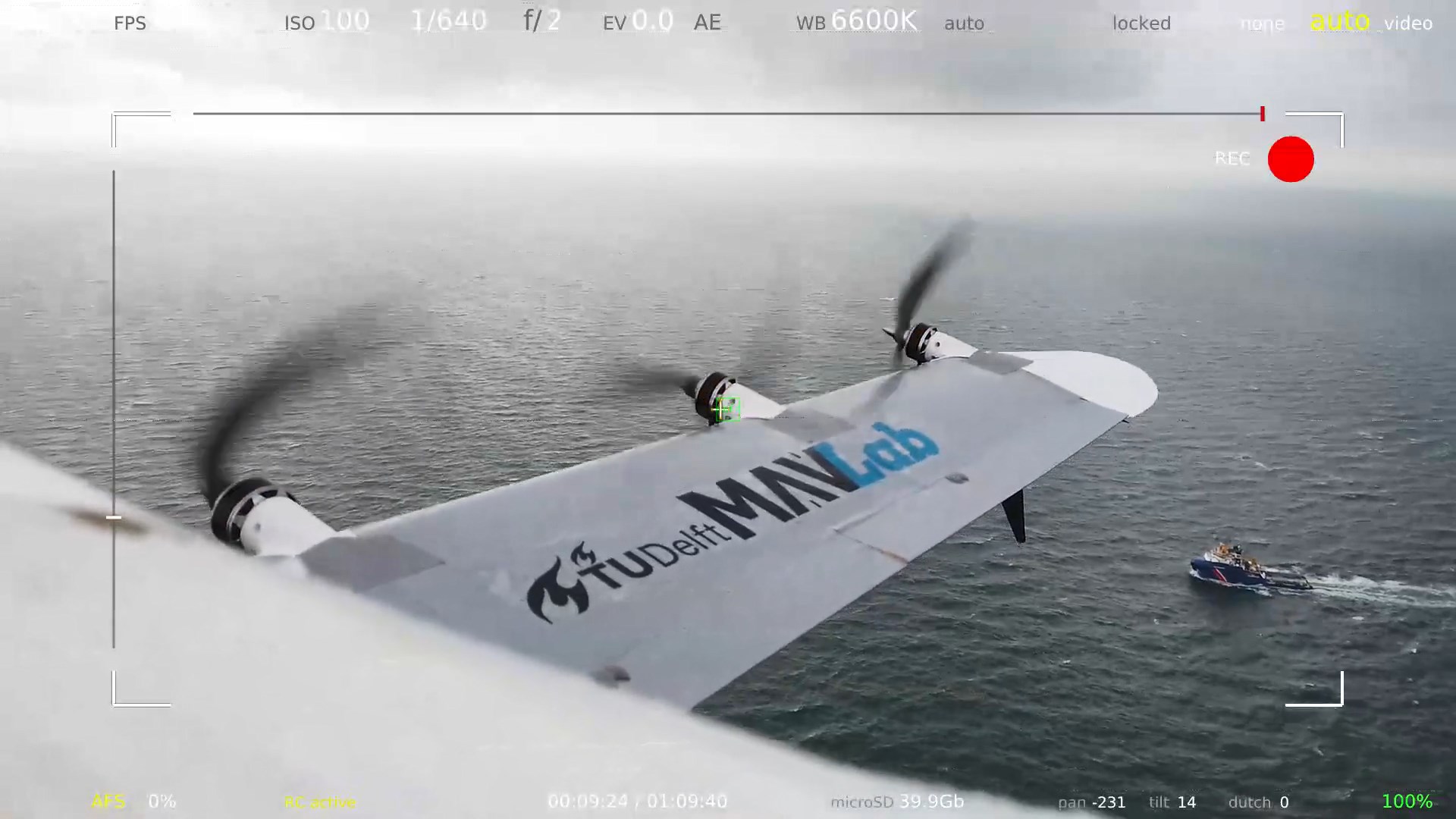}
\caption{Live-view of the telemetry during the long hydrogen flight test at sea.}\label{fig:flight:onboard}
\end{figure}

The onboard video of the pan-tilt HD camera protruding from the top of the \nederdrone was streamed via the 2.4GHz ISM-band HereLink data-link.
A live video view of the \nederdrone following the ship is shown in \fig\ref{fig:flight:onboard}.
All battery powered data-links and video systems were also charging from the hydrogen energy and stayed fully charged during the entire flight.

After landing, the empty cylinder can be replaced with a new full cylinder in seconds, before taking off again.
The presented test flight does not push endurance to its limits.
There was at least 20 minutes worth of hydrogen and 15 minutes of battery left after landing.
All systems were running at full power and the weather was rough with 5 Beaufort (20kt) wind and moderate turbulence.
The propellers used during this flight were optimized for fast flight and not for maximum endurance.
This illustrates that the hybrid lift hybrid power \ac{UAV} called \nederdrone is built for real world operations and has considerable safety, performance and energy margins.

\subsection{Energy profile of the 3h flight}

\begin{figure}[hbt]
\centering
\makebox[0.99\linewidth][c]{\includegraphics[scale=0.8]{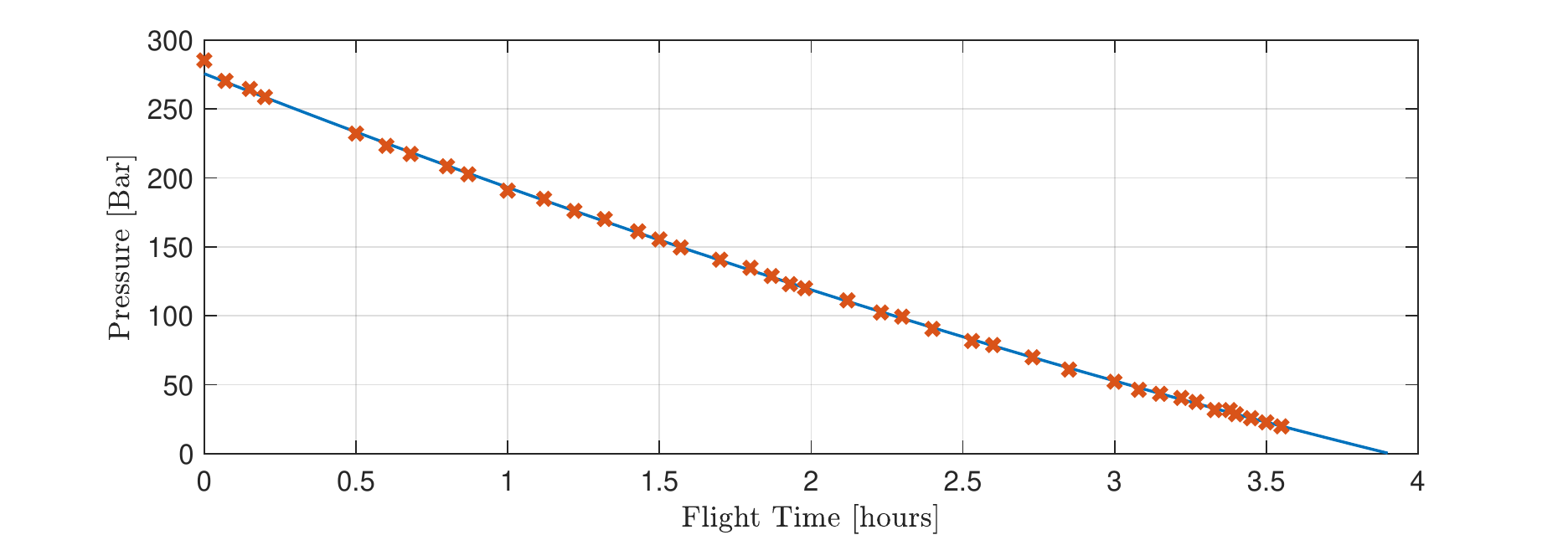}}
\caption{Depletion of the cylinder in function of flight time.} \label{fig:flight:percent}
\end{figure}

\begin{figure}[hbt]
\centering
\makebox[0.99\linewidth][c]{\includegraphics[scale=0.8]{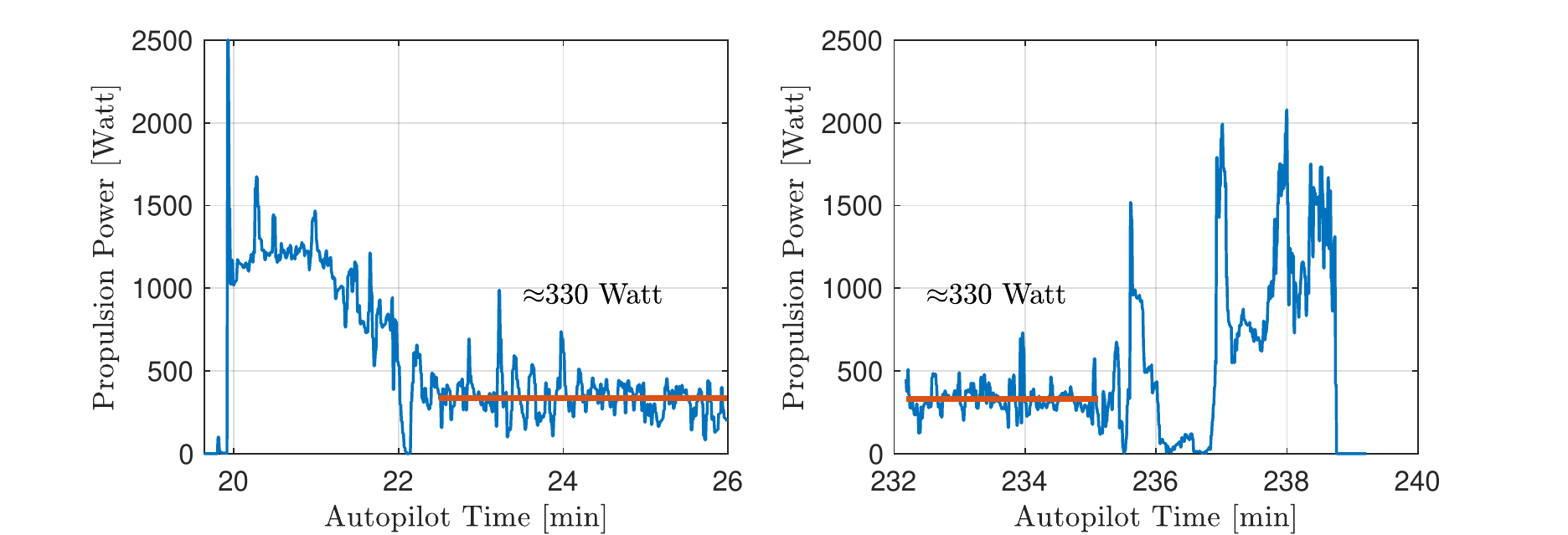}}
\caption{Flight power as reported by the \ac{ESC} during the take-off and landing phases.} \label{fig:flight:power}
\end{figure}

\begin{figure}[hbt]
\centering
\makebox[0.99\linewidth][c]{\includegraphics[scale=0.8]{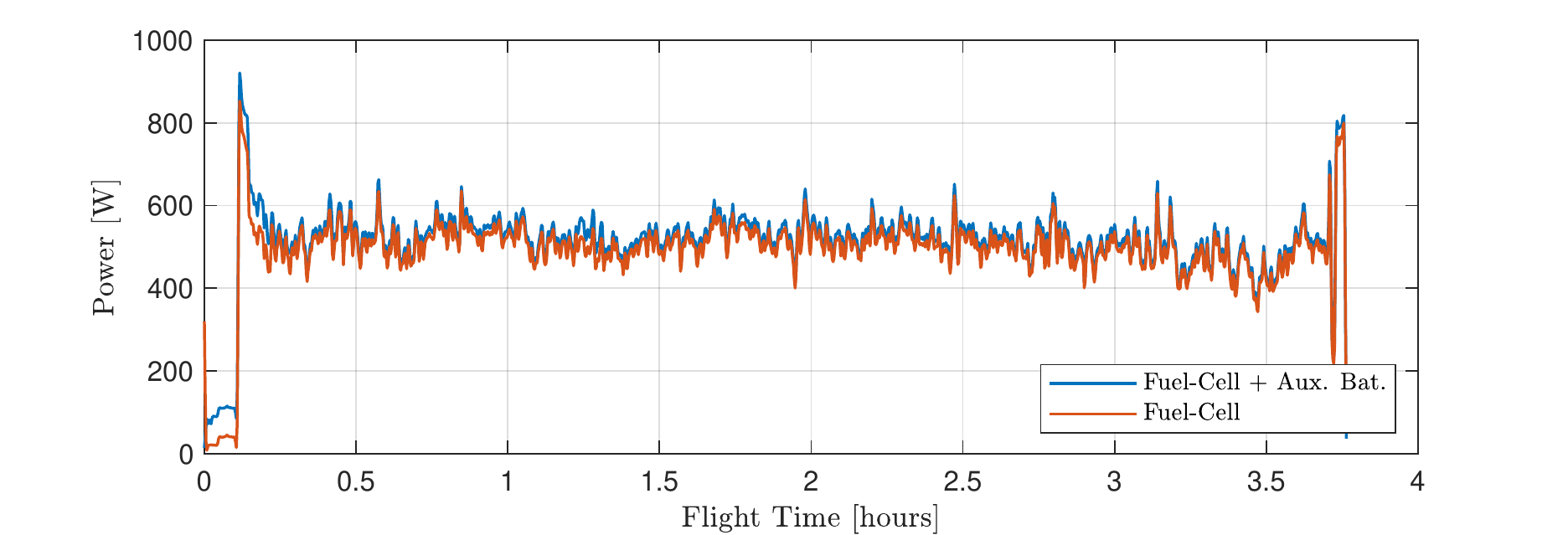}}
\caption{Power generation of the IE800 fuel-cell.} \label{fig:flight:power:fuelcell}
\end{figure}

The hydrogen cylinder was filled with a pressure of \SI{285}{\bar} after settling at ambient temperature.
\fig\ref{fig:flight:percent} shows the depletion of the hydrogen cylinder as measured by the onboard sensors.
It follows the inverse of the density profile from \eqref{eq:hydrogen}.
Having reached the desired 3 flight hours, the flight was stopped at a remaining pressure of \SI{20}{\bar}, although previous tests proved that the \nederdrone can continue to fly safely on battery power after complete depletion of the cylinder and shutdown of the fuel-cell.

During take-off and landing, lithium batteries provide the required extra power while the fuel-cell is running at maximum power.
\fig\ref{fig:flight:power} shows the reported power used by the \ac{ESC}s.
This excludes the power used by the fuel-cell itself and its cooling, power losses in the long wires, power loss over the diode,
and power used by the payload and video link.
The descend power becomes nearly zero at moments when the \nederdrone is 
gliding in forward flight with the propellers windmilling.
The climb power (from 20 to \SI{21}{\minute}) is about \SI{1250}{\watt} during the angled take-off with the wings not stalled and thus significantly helping in lift production.
The hover power required in the last phase of the landing while fighting turbulence with the wing stalled (\SI{238}{\minute}) consumed nearly \SI{1500}{\watt} with peaks of over \SI{2000}{\watt}.
This is much more than the raw fuel-cell system can handle but is supplied from the high current rated lithium hover batteries with ease. 
\fig\ref{fig:flight:power:fuelcell} shows that the power delivered by the fuel-cell in those cases is about \SI{800}{\watt} as by design.
Two minutes after the take-off, the \nederdrone transitions to forward flight and starts using much less power.
The sum of the flight power, payload power and fuel-cell systems (including cooling) power are an average of \SI{550}{\watt}.
After the take-off the fuel-cell slowly re-charges the lithium batteries that were used during take-off.
This means the hovering lithium batteries are fully charged before landing.

\section{Discussion}\label{sec:discussion}

Hydrogen is seen as a highly promising future fuel for aviation thanks to its high power density.
But the limited power that can be generated by fuel-cells limits the applicability.
Furthermore, the onboard storage of pure hydrogen requires a pressure cylinder with a weight that is easily one quarter of the vehicle weight and has shape constraints.

To allow the successful application of hydrogen in UAVs, it is important that the vehicle does not have severe operational limitations and it is primordial that the safety is guaranteed.
This underlines the importance to find concepts that do not need very long runways but nevertheless fly fast and efficiently.
At the same time these platforms must be very safe as the consequences of accidents with onboard pressure cylinders filled with hydrogen can be significant.
This requires platforms with redundant flight modes, redundant energy and redundant control.
The shape of the \ac{UAV} can also play a big role in the protection of the cylinder.
Light foam around the cylinder provides both an aerodynamic shape and a large crumble zone for a low weight.
By placing the sensitive high-pressure regulators backwards in the middle of the vehicle, safety can be further increased.
Last but not least, by having dual flight modes, an additional recovery mode is created in case of failure.
When for instance aerodynamic actuators would fail, then the platform can return and land in hovering flight.
If on the other hand many motor controllers would fail,
then the platform can still be flown in forward flight by exploiting the efficiency of its fixed-wings.
This combined versatility and safety is expected to play an important role in the development of hydrogen fuelled flight.

\section{Conclusions}\label{sec:conclusion} 

A novel hydrogen \ac{UAV} was presented called the \nederdrone\footnote{\myurl{http://www.nederdrone.nl/}}.
It is a tail-sitter hybrid lift vehicle with tandem wings for forward flight, and 12 propellers for hover.
The power comes from a \ac{PEM} fuel-cell with hydrogen stored in a pressurized cylinder around which the \ac{UAV} is optimized.
The dual automotive CAN control bus, redundant power source, wiring, propulsion, dual flight modes and model-less INDI control make the \nederdrone particularly resilient to failures.
The versatility and flight endurance of the \nederdrone is shown with a 3h38 test flight at sea from a moving ship with \SI{20}{kt} winds.

\appendix


\section*{Acknowledgements}
The developments presented in this work would not have been possible without the support from the Royal Netherlands Navy and the Netherlands Coastguard.


\section{Cylinder Overview}

An overview of the considered pressurized hydrogen cylinders is given in Table~\ref{table:tanks}.

\begin{table*}[htb]
\begin{center}
\begin{tabular}{|llllllllll|}
\hline
	&	$V$ &	$p$	&	$E$	&	$W$	&	H$_2$	&	[Wh	&	WT\%	&	$D$ 	&	$L$ 	\\
	&	[L]	&	[bar]	&	[Wh]	&	 [kg]	&	[g]	&	/kg]	&	H$_2$	&	 [mm]	&	 [mm]	\\ \hline
HES 	&	2	&	350	&	1564	&	1.2	&	46.96	&	1303	&	3.91\%	&	102	&	385	\\
A-Series	&	2.5	&	350	&	1955	&	1.25	&	58.7	&	1564	&	4.70\%	&	132	&	228	\\
	&	3.5	&	350	&	2737	&	1.65	&	82.2	&	1659	&	4.98\%	&	132	&	375	\\
	&	5	&	350	&	3910	&	1.85	&	117.4	&	2113	&	4.44\%	&	152	&	395	\\
	&	9	&	350	&	7037	&	2.85	&	211.3	&	2469	&	7.41\%	&	173	&	528	\\
	&	12	&	350	&	9383	&	3.5	&	281.8	&	2681	&	8.05\%	&	196	&	532	\\
	&	20	&	350	&	15638	&	7	&	469.6	&	2234	&	6.71\%	&	230	&	665	\\
\hline
Luxfer	&	3.8	&	379	&	3173	&	2.5	&	95.3	&	1269	&	3.81\%	&		&		\\
	&	5.7	&	379	&	4759	&	3.3	&	142.9	&	1442	&	4.33\%	&		&		\\
	&	6.8	&	300	&	4671	&	3.3	&	140.3	&	1415	&	4.25\%	&	158	&	520	\\
   	&	7.6	&	379	&	6345	&	4.1	&	190.6	&	1548	&	4.65\%	&		&		\\
 	&	9	&	300	&	6182	&	4.3	&	185.7	&	1438	&	4.32\%	&		&		\\
\hline
CTS	&	2	&	300	&	1374	&	1.23	&	41.3	&	1118	&	3.36\%	&		&		\\
	&	3	&	300	&	2061	&	1.6	&	61.9	&	1288	&	3.87\%	&		&		\\
	&	6	&	300	&	4121	&	2.9	&	123.8	&	1421	&	4.27\%	&		&		\\
	&	6.8	&	300	&	4671	&	3.1	&	140.3	&	1507	&	4.52\%	&	161	&	520	\\
	&	7.2	&	300	&	4946	&	3.3	&	148.5	&	1499	&	4.50\%	&	161	&	545	\\
	&	9	&	300	&	6182	&	4.3	&	185.6	&	1438	&	4.32\%	&		&		\\
	&	13	&	300	&	8930	&	5.3	&	268.2	&	1685	&	5.06\%	&		&		\\
\hline
HES 	&	2	&	300	&	1374	&	1.2	&	41.26	&	1145	&	3.44\%	&	113	&	369	\\
F-Series	&	3	&	300	&	2061	&	1.4	&	61.9	&	1472	&	4.42\%	&	122	&	440	\\
	&	6	&	300	&	4121	&	2.5	&	123.8	&	1649	&	4.95\%	&	161	&	481	\\
	&	6.8	&	300	&	4671	&	2.7	&	140.3	&	1730	&	5.20\%	&	161	&	520	\\
	&	7.2	&	300	&	4946	&	2.8	&	148.5	&	1766	&	5.30\%	&	161	&	545	\\
	&	9	&	300	&	6182	&	3.8	&	185.6	&	1627	&	4.89\%	&	182	&	543	\\
\hline
\end{tabular}
\end{center}
\caption{Cylinder Overview. Volume $V$, maximum pressure $p$, energy content $E$, weight $W$ (regulator not included), hydrogen weight, specific weight percent of hydrogen, diameter $D$ and length $L$}\label{table:tanks}
\end{table*}


\bibliographystyle{elsarticle-num}
\bibliography{literature}

\end{document}